\pdfoutput=1

\documentclass[11pt]{article}

\usepackage[]{acl}

\usepackage{times}
\usepackage{latexsym}

\usepackage[T1]{fontenc}

\usepackage[utf8]{inputenc}

\usepackage{microtype}

\usepackage{subcaption}
\usepackage{multirow}
\usepackage{pifont}
\usepackage{amssymb}
\usepackage{bm}
\usepackage{amsmath}
\usepackage{booktabs}
\usepackage{makecell}
\usepackage{colortbl}
\usepackage{array}
\usepackage{graphicx}
\usepackage[figuresright]{rotating}
\usepackage{xspace} 
\usepackage{bbold}
\usepackage{tabularx}
\usepackage{fontawesome}
\usepackage{comment}

\usepackage{CJKutf8}

\definecolor{pinegreen}{RGB}{15,153,15}
\definecolor{mygray}{gray}{.9}
\definecolor{target}{RGB}{0,0,146}
\definecolor{myblue}{RGB}{25,101,255}
\definecolor{myorange}{RGB}{239,134,63}

\newcommand*{\circled}[1]{\lower.7ex\hbox{\tikz\draw (0pt, 0pt)%
		circle (.5em) node {\makebox[1em][c]{\small #1}};}}

\definecolor{bestd}{RGB}{237,100,152}
\definecolor{bestc}{RGB}{0,126,219}

\newcommand{\myroman}[1]{\uppercase\expandafter{\romannumeral#1}}

\makeatletter
\newcommand*\bigcdot{\mathpalette\bigcdot@{.5}}
\newcommand*\bigcdot@[2]{\mathbin{\vcenter{\hbox{\scalebox{#2}{$\m@th#1\bullet$}}}}}
\makeatother

\newcolumntype{P}[1]{>{\centering\arraybackslash}p{#1}}

\newcommand{\logofig}{\includegraphics[height=0.7em,trim=0 0em 0 0]{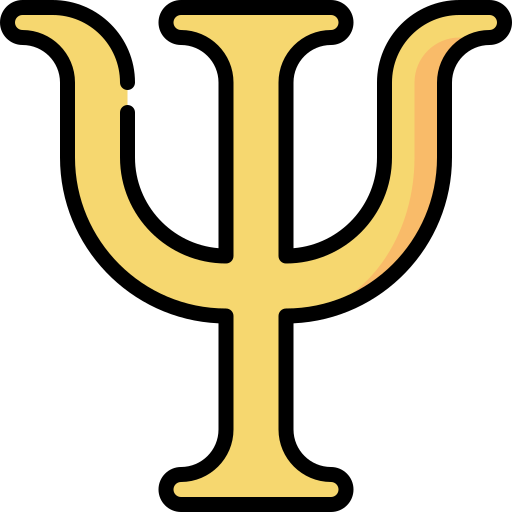}}
\newcommand{\logo}{\logofig-\textsc{Arena}\xspace}

\newcolumntype{\myline}{!{\vrule width 0.08em}}

\title{\logo: Interactive Assessment and Optimization of LLM-based Psychological Counselors with Tripartite Feedback}

\author{
 \textbf{Shijing Zhu\textsuperscript{1}},
 \textbf{Zhuang Chen\textsuperscript{1,2}},
 \textbf{Guanqun Bi\textsuperscript{2}},
 \textbf{Binghang Li\textsuperscript{3}},
\\
 \textbf{Yaxi Deng\textsuperscript{1}},
 \textbf{Dazhen Wan\textsuperscript{3}},
 \textbf{Libiao Peng\textsuperscript{3}},
 \textbf{Xiyao Xiao\textsuperscript{3}},
\\
 \textbf{Rongsheng Zhang\textsuperscript{4}},
 \textbf{Tangjie Lv\textsuperscript{4}},
 \textbf{Zhipeng Hu\textsuperscript{4}},
 \textbf{FangFang Li\textsuperscript{1}},
\textbf{Minlie Huang\textsuperscript{2}}
\\
\\
 \textsuperscript{1}Central South University,
 \textsuperscript{2}CoAI Group, DCST, IAI, BNRIST, Tsinghua University,\\
 \textsuperscript{3}Lingxin AI,
 \textsuperscript{4}Fuxi AI Lab, NetEase Inc.
\\
{zhchen18@foxmail.com}}

\begin{document}
\maketitle
\begin{abstract}
Large language models (LLMs) have shown promise in providing scalable mental health support, while evaluating their counseling capability remains crucial to ensure both efficacy and safety. Existing evaluations are limited by the static assessment that focuses on knowledge tests, the single perspective that centers on user experience, and the open-loop framework that lacks actionable feedback. To address these issues, we propose \logo, an interactive framework for comprehensive assessment and optimization of LLM-based counselors, featuring three key characteristics: (1) Realistic arena interactions that simulate real-world counseling through multi-stage dialogues with psychologically profiled NPC clients; (2) Tripartite evaluation that integrates assessments from the client, supervisor, and counselor perspectives; (3) Closed-loop optimization that iteratively improves LLM counselors using diagnostic feedback. Experiments across eight state-of-the-art LLMs show significant performance variations in different real-world scenarios and evaluation perspectives. Moreover, reflection-based optimization results in up to a 141\% improvement in counseling performance. We hope \logo provides a foundational resource for advancing reliable and human-aligned LLM applications in mental healthcare.
\end{abstract}

\section{Introduction}

Mental health disorders affect over 1 billion people globally, with the World Health Organization noting their significant societal and economic impacts, such as reduced productivity and strained healthcare systems \cite{WHO2023mentalhealth}. However, there is a severe shortage of mental health professionals, with approximately 100,000 people per counselor. This shortage has driven the exploration of AI-based counseling systems as a potential solution. In the 1960s, early rule-based AI systems like ELIZA \cite{weizenbaum1966eliza} showed the feasibility of automated counseling. Today, large language models (LLMs) like GPT-4 \cite{achiam2023gpt} and Claude \cite{Anthropic2023Claude} exceed human abilities in certain tasks, prompting increasing efforts to use LLMs for scalable counseling and make mental services more accessible \cite{chen2023llm,iftikhar2024therapy,xu2025autocbt}. This trend highlights the urgent need for rigorous evaluation to ensure that these systems meet clinical standards for effectiveness, control, and safety.

\begin{figure}[t]
	\centering
	\vspace{0mm}
	\hspace{-4mm}
	\setlength{\abovecaptionskip}{2mm}
	\setlength{\belowcaptionskip}{0mm}
	\includegraphics[width=0.49\textwidth]{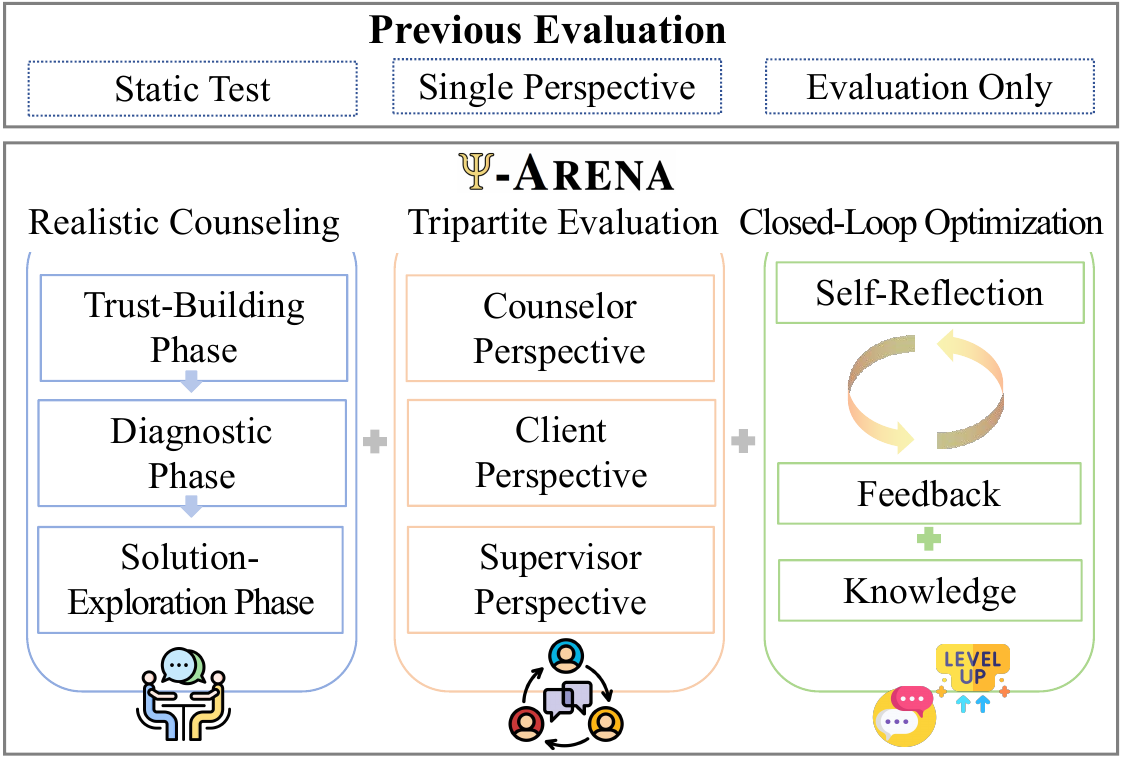}
	\centering
	\caption{The comparison between \logo and existing studies on evaluating LLM-base counselors.}
	\label{fig-intro}
	\vspace{-2mm}
\end{figure}

 
Although pioneering studies have attempted to evaluate LLM counselors, three key challenges remain that prevent comprehensive and in-depth assessments: \textbf{1) Gap between understanding and application}. Existing studies tend to focus on static assessments, such as multiple-choice questions or diagnostic accuracy metrics, which measure knowledge rather than practical application \cite{jin2023psyeval,zhang2024cbt}. \textbf{2) Limited user-centric metrics}. While \citet{zhao2024esc} and \citet{wang2024towards} try to simulate counseling interactions between clients and counselors, they primarily focus on client satisfaction and subjective feelings, ignoring evaluations from supervisors and counselors themselves. \textbf{3) Lack of feedback loops}. Most existing frameworks lack actionable feedback for model improvement, which should be a key objective of any evaluation system. 


In this paper, we propose \logo, an interactive platform for assessing and optimizing LLM-based psychological counselors. In \logo, LLM counselors engage with virtual NPC clients and receive assessments from three perspectives: the client, the supervisor, and the counselor. These evaluations provide targeted feedback that helps optimize the counseling process. Specifically, \logo features three key elements: \textbf{1) Realistic counseling scenarios}. To ensure the arena simulates real-world counseling, we focus on client profiles and behaviors. For profiles, we analyze real counseling records to identify key attributes for virtual clients' psychological profiles and create 10,000 virtual client profiles (NPC cards) across 100 topics for use. For behaviors, we base the simulation on professional counseling knowledge, ensuring meaningful interactions across different phases: ``\textit{trust-building}$ \rightarrow$ \textit{diagnosis} $\rightarrow$ \textit{solution exploration}''. \textbf{2) Tripartite evaluation metrics}. We integrate evaluations from clients (subjective experience), supervisors (professional competency), and counselors (reflective awareness), enabling a 360° competency analysis across 33 dimensions. \textbf{3) Closed-loop optimization}. We introduce a feedback and optimization cycle, where evaluation results are combined with professional counseling guides to generate specific feedback, allowing LLM counselors to self-reflect and iteratively improve their responses.

In \logo, we evaluate eight state-of-the-art LLMs, including closed-source models like Claude-3.5-Sonnet and open-source models like DeepSeek-671B. Our results show significant performance disparities across these LLM counselors when evaluated from different perspectives, emphasizing the need for arena simulations and multi-source evaluations. We also compare the automatic evaluation results with those of human experts, revealing high consistency and validate the effectiveness. Additionally, through specific feedback and optimization, we achieve up to a 141\% improvement in counseling performance, showcasing the potential of a closed-loop evaluation system.


Our key contributions are: (1) Introducing \logo, which features realistic counseling scenarios, tripartite evaluation metrics, and closed-loop optimization. (2) Evaluating the counseling performance of state-of-the-art LLMs and demonstrating consistency with human experts, achieving performance improvements based on feedback. (3) Conducting in-depth analysis of LLM performance across various dimensions and topics. We hope \logo to serve as an efficient and effective evaluation framework, advancing the responsible development of LLMs in mental healthcare.


\begin{figure*}[t]
	\centering
	\vspace{0mm}
	\hspace{-4mm}
	\setlength{\abovecaptionskip}{2mm}
	\setlength{\belowcaptionskip}{0mm}
	\includegraphics[width=0.99\textwidth]{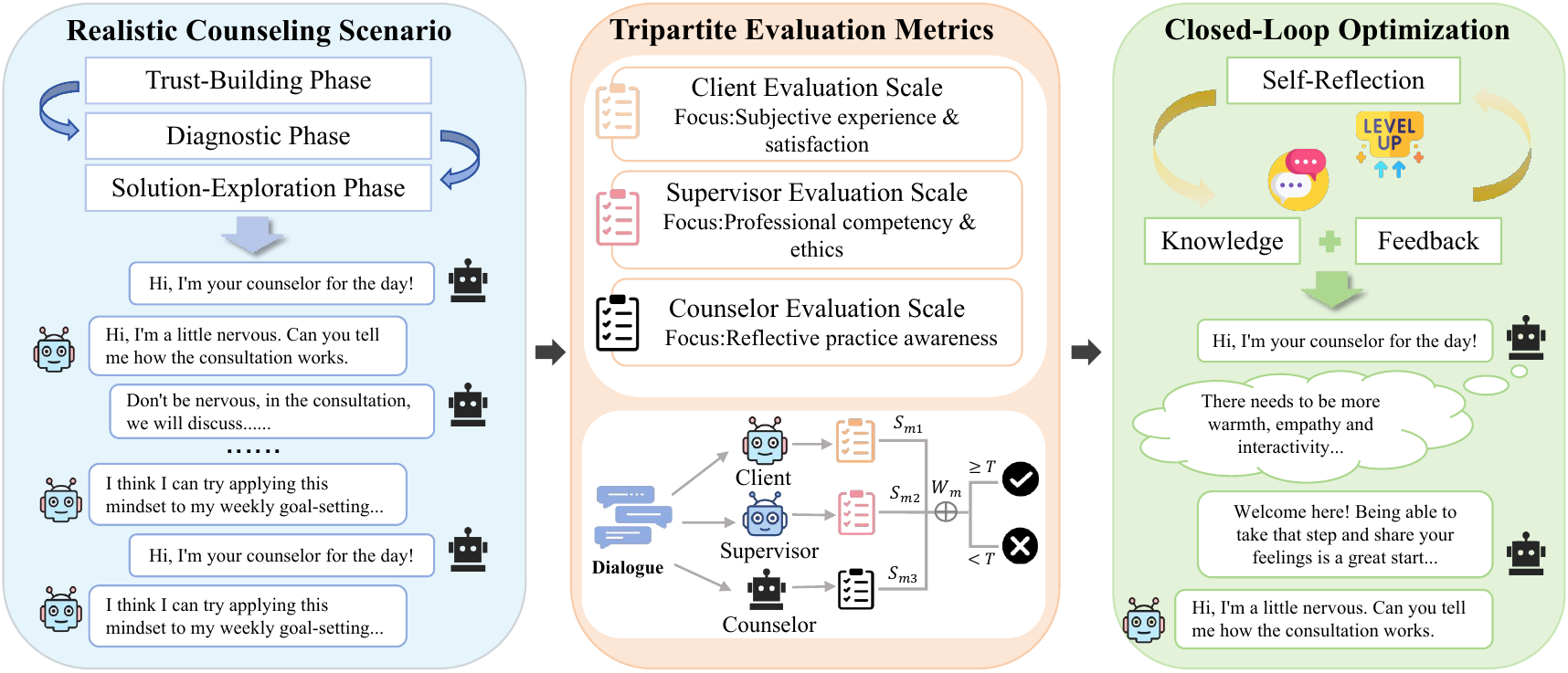}
	\centering
	\caption{In \logo, LLM-based counselors interact with NPC clients, receive multi-source evaluations, and improve counseling performance through self-reflection.}
	\label{fig-framework}
	\vspace{-2mm}
\end{figure*}

\section{\logo}
\subsection{Framework Overview}
As shown in Figure \ref{fig-framework}, \logo is an interactive framework for assessing and optimizing LLM-based psychological counselors. \logo encompasses virtual clients with diverse psychological profiles who engage in multi-stage counseling dialogues with LLM counselors. Then \logo evaluates counselor performance from three perspectives: client, supervisor, and counselor. Based on these evaluations, \logo generates feedback to guide the counselor’s self-reflection and iterative improvement.

\subsection{Client in \logo}
In \logo, virtual NPC clients are created with rich psychological profiles and realistic behaviors to ensure that the simulation of counseling scenarios is both diverse and authentic. 
\subsubsection{Client Profiles}
The construction of profiles is based on real-world counseling records, ensuring that the virtual clients reflect authentic psychological concerns \cite{DZXI202007003}. Each profile includes several key attributes that define the client’s background, emotional state, and the issues they seek counseling for. The below attributes are extracted and incorporated into the client profiles: \textit{demographics}, \textit{cultural background}, \textit{personality trait}, \textit{emotional state}, \textit{current distress}, \textit{detailed distress description}, and \textit{core theme}. {Details of attributes can be found in Appendix \ref{app:profile}}.


To ensure high-quality and diverse client profiles, we resort to the real-world PsyQA dataset \cite{psyqa} which contains conversations from real clients, covering common mental health disorders across nine themes, including self-growth, emotional issues, relationships, behavior, family, therapy, marriage, and career. Each theme contains several subtopics, ultimately generating 100 distinct topics for client profiles. {Details of topics can be found in Appendix \ref{app:topics}}.
To build these profiles, we use GPT-4o to extract initial psychological profiles from PsyQA. The extraction process is guided by carefully designed prompts to capture each profile’s key attributes. {Detailed prompts can be found in Appendix \ref{app:prompt}}. We create 10,000 high-quality client profiles, and then, for each topic, we manually select one high-quality profile to serve as the basis for constructing the NPC clients. 

\subsubsection{Client Behaviors}

To simulate realistic counseling interactions, we design client behaviors that match the different stages of a typical counseling process. These behaviors help ensure the virtual clients engage in meaningful conversations with LLM counselors. Based on counseling models from existing research, we focus on three main phases of client behavior during the simulation. 

\paragraph{Trust-Building Phase}{\cite{Sachse2024}} In the beginning, the virtual client works on building trust by being open to the counselor’s questions, sharing personal feelings, and offering context about their struggles. The client might show vulnerability, helping establish a connection and encourage a safe space for further discussions. 
\paragraph{Diagnostic Phase}{\cite{zhang2024cpsycoun}} During this phase, the client begins to share more personal information, such as their background, emotional state, and the deeper causes of their distress. They may reflect on past experiences and feelings, providing the counselor with insights into what might be influencing their current struggles. 
\paragraph{Solution-Exploration Phase}{\cite{hill2020helping}} In the final phase, the client actively explores possible solutions and coping strategies. They may express hope, consider different options, or reflect on past efforts to resolve their issues. The client may also ask for advice and discuss potential next steps for moving forward.

\subsubsection{Client Simulation}

To ensure authenticity, the client's responses are simultaneously guided by their defined psychological profiles and behavior patterns. To align the virtual client's dialogue with real counseling scenarios, we follow five core principles when instructing GPT-4o for client simulation {\cite{tu2024charactereval,shao2023character,chen2024persona}}: realism (ensuring the conversation matches the client's language style and emotional expression), fluency (maintaining logical and natural dialogue flow), completeness (covering key tasks across all counseling stages), personalization (reflecting the client's unique background and traits), and behavioral consistency (ensuring stable behavior patterns across different conversation turns). {Detailed prompts for client simulation can be found in Appendix \ref{app:prompt}}.

\subsection{Counselor in \logo}


\begin{table*}[ht]
\vspace{0mm}
\small
\renewcommand\arraystretch{1.2}
\setlength{\abovecaptionskip}{0mm}
\setlength{\belowcaptionskip}{5mm}
\centering
\setlength{\tabcolsep}{0.8mm}
\begin{tabular}{l|l|l|l}
\toprule
\textbf{Perspective}        & \textbf{Evaluation Focus}             & \textbf{Scale Characteristics}               & \textbf{Realistic Threshold}           \\ \midrule
Client               & Subjective experience \& satisfaction & 16-dimension scale (0-4 per item)            & >42 (Total 64)              \\ \midrule          
Supervisor           & Professional competency \& ethics    & 8-dimension APA scale (0-4 per item)         & >24 (Total 32)  
\\ \midrule     

Counselor       & Reflective practice awareness         & 9-dimension ability scale (0-5 per item)     & >35 (Total 45) \\ \bottomrule
\end{tabular}

\vspace{1mm}
\caption{Evaluation criteria and thresholds of tripartite scales. 
}

\label{tab-scale}
\vspace{-6mm}
\end{table*}

In \logo, the evaluated LLMs act as counselors and engage in conversations with different clients. To ensure a fair evaluation of LLM-based counselors, we provide each model with a standardized \texttt{psycho prompt} that clearly defines its role as a psychological counselor. The prompt also outlines the basic structure of the counseling session, including the key phases, interaction rounds, and the overall session length. No additional guiding information is given to avoid artificially boosting performance. {Further details of counselor prompts are provided in Appendix \ref{app:prompt}}. For comparison, we also include the default \texttt{system prompt} (e.g., ``you are a helpful assistant'') to observe the vanilla performance without any specific instruction.

\subsection{Tripartite Evaluation Metrics}

To ensure a comprehensive evaluation of LLM-based psychological counselors, we introduce a tripartite evaluation system that assesses the counseling dialogue from three distinct perspectives: the client, the supervisor, and the counselor. This approach, inspired by established frameworks in psychology \cite{lockyer2003multisource,kuzmits2004360,tham2007360}, provides a holistic assessment of the counselor’s professional abilities, as shown in Table \ref{tab-scale}. Below, we briefly describe each evaluation scale and its core focus areas. {The detailed scoring items are available in the Appendix \ref{app:scales}}.

\paragraph{Client-Oriented Scale} The client-oriented scale, developed by Joel Black in ``\textit{Who Stole Your Trust and Confidence}''? {\cite{trust}}, gathers feedback from clients on the counselor’s effectiveness and the quality of the counselor-client relationship. It includes 16 dimensions such as trust, empathy, and communication clarity, rated on a scale from 0 to 4. The focus is on the client’s emotional experience and satisfaction with the counselor’s approach, offering insights into the counseling process.

\paragraph{Supervisor-Oriented Scale} The Supervisor-Oriented Scale, based on the ``\textit{Consultant Competency Assessment Tool}'' developed by the American Psychological Association \cite{apa2023website}, evaluates the counselor’s professional competence and adherence to ethical standards. It includes 8 dimensions, such as therapeutic techniques and cultural competence, rated from 0 to 4. This scale emphasizes the counselor’s technical skills, ethical conduct, and adherence to professional practices.

\paragraph{Counselor-Oriented Scale} The counselor-oriented scale, developed by \citet{ZGXG201811001617}, allows counselors to self-assess their practices. It covers 20 dimensions, with 9 focused on practical counseling abilities like empathy and client response. Counselors rate their alignment with these abilities on a scale from 0 to 5. This scale encourages self-awareness and supports ongoing professional development.

For each evaluation session, we use GPT-4o to simulate the roles of the client, supervisor, and counselor, following the evaluation scales mentioned above and assigning appropriate scores. {Detailed prompts for these role-playing scenarios are provided in Appendix \ref{app:prompt}}. To validate the effectiveness of the automatic scoring system, we also recruit psychological experts to score a sample of 30 sessions conducted by various LLM counselors. We then compare the consistency between the automatic evaluations and the human expert scores. The results of this comparison are presented in Section \ref{sec:results} to demonstrate the reliability and effectiveness of the tripartite evaluation system.

\subsection{Closed-Loop Optimization}

\begin{table*}[ht]
	\vspace{0mm}
	\small
	\renewcommand\arraystretch{1.6}
	\setlength{\abovecaptionskip}{0mm}
	\setlength{\belowcaptionskip}{5mm}
	\centering
	\setlength{\tabcolsep}{0.8mm}
		\resizebox{1.\textwidth}{!}{
\begin{tabular}{c|cccc|cccc|c}
\toprule
	\multirow{2}{*}{Model} & \multicolumn{4}{c|}{\texttt{Vanilla Prompt}}                                & \multicolumn{4}{c|}{\texttt{Psycho Prompt}}                                 & {$\delta$ } \\
	& Client & Supervisor & Counselor & Overall & Client & Supervisor & Counselor & Overall & Pass Rate                       \\ \midrule
	GPT-3.5 Turbo          & 2.68 (57\%)           & 2.60 (13\%)               & 3.43 (10\%)              & 2.90 (33\%)           & 2.61 (50\%)           & 2.66 (20\%)               & 3.60 (22\%)              & 2.96 (39\%)            & +6\%                      \\
	GLM-4-Plus             & 2.58 (47\%)           & 2.41 (9\%)                & 3.32 (12\%)              & 2.77 (16\%)            & 2.53 (43\%)           & 2.64 (14\%)               & 3.71 (38\%)              & 2.96 (39\%)            & +23\%                     \\
	MiniMax-Text-01        & 2.61 (56\%)           & \textbf{2.71} \colorbox{yellow}{(23\%)}               & \textbf{3.59} (19\%)              & 2.97 (35\%)            & 2.64 (60\%)           & 2.77 (27\%)               & 3.62 (25\%)              & 3.01 (46\%)            & +11\%                     \\
	LLaMA-3.3              & 2.47 (21\%)           & 2.53 (9\%)                & 3.46 (17\%)              & 2.82 (15\%)            & 2.63 (47\%)           & 2.65 (18\%)               & 3.78 (45\%)              & 3.02 (48\%)            & +33\%                     \\
	GPT-4o                 & \textbf{2.73} \colorbox{yellow}{(73\%)}           & 2.69 (17\%)               & 3.54 (19\%)              & \textbf{2.99} \colorbox{yellow}{(44\%)}            & 2.69 (64\%)           & 2.80 (30\%)               & 3.78 (42\%)              & 3.09 (57\%)            & +13\%                     \\
	Qwen-2.5               & 2.62 (64\%)           & 2.68 (15\%)               & 3.56 \colorbox{yellow}{(22\%)}              & 2.95 (33\%)            & 2.71 (65\%)           & 2.74 (26\%)               & 3.72 (37\%)              & 3.06 (59\%)            & +26\%                     \\
	Deepseek               & 2.58 (53\%)           & 2.60 (9\%)                & 3.43 (9\%)               & 2.87 (28\%)            & 2.64 (58\%)           & 2.81 (34\%)               & 4.03 (60\%)              & 3.16 (65\%)            & +37\%                     \\
	Claude 3.5 Sonnet      & 2.66 (59\%)           & 2.65 (17\%)               & 3.53 (20\%)              & 2.95 (37\%)            & \textbf{2.75} \colorbox{yellow}{(72\%)}           & \textbf{2.87} \colorbox{yellow}{(47\%)}               & \textbf{4.08} \colorbox{yellow}{(77\%)}              & \textbf{3.23} \colorbox{yellow}{(84\%)}            & \colorbox{yellow}{+47\%}      \\ \bottomrule              
\end{tabular}
		}
		
		\vspace{1mm}
		\caption{Evaluation results of \logo of all LLMs including both scores (best in bold) and pass rates (best in yellow). The score reflects the model's average performance across all tests, while the pass rate measures the number of passed tests. Therefore, the highest score and the highest pass rate may be achieved by different models.
		}
		
		\label{tab-main-results}
		\vspace{-6mm}
	\end{table*}
The optimization process in \logo aims to improve LLM-based counselors through iterative self-reflection. Based on the evaluation results, we use GPT-4o to automatically generate feedback for low-scoring responses. Specifically, we draw from established psychological frameworks, such as "\textit{Practice of Counseling and Psychotherapy}" \cite{corey2013theory}, to infuse the LLM with knowledge from 11 major counseling approaches. These include cognitive-behavioral therapy (CBT), person-centered therapy, and psychodynamic approaches. We instruct the LLM to provide model-specific feedback after each session.

After receiving the feedback, the optimization process follows three steps for each counseling turn. First, the LLM counselor reviews its previous response and reflects on its strengths and weaknesses based on the feedback. Next, the counselor rewrites the response to improve it. Finally, the new version of the response is re-evaluated. By comparing the scores before and after the revision, we assess whether \logo can help counselors improve in a way similar to real-world supervision. This comparison data has the potential to be used in subsequent model post-training to enhance generalization abilities. Detailed prompts for feedback generation and self-reflection are provided in Appendix \ref{app:prompt}.

\section{Experiment}
\subsection{Settings \& Metrics}


In the experiment, each LLM-based counselor engages in individual dialogues with 100 virtual clients. Each counseling session consists of 25 conversational rounds. The dialogue content is then evaluated using a tripartite scoring system, and feedback is generated to improve the model.

For metrics, we calculate the average scores on each of the three scales and determine an overall mean score. We also introduce the "pass rate," a metric commonly used in real-world counselor assessments, to provide a clear view of counseling performance. Specifically, each scale has a threshold to determine whether the counselor meets the required standards: the client scale (Total > 42), the counselor scale (Total > 35), and the supervisor scale (Total > 24). We analyze the pass rate on each scale and also compute the overall pass rate, which considers a counselor as passing only if they meet the thresholds on all three scales.

To validate the effectiveness of the automated evaluation, we also recruit two graduate students with a background in psychology to manually label the model's performance. These two experts are very familiar with the research content, and we provide them with the complete tripartite scales and necessary instructions to ensure they are fully equipped to carry out the labeling task. We pay each expert an hourly rate of \$13.78.

\subsection{Models}
We evaluate eight state-of-the-art large language models. The closed-source models selected are GPT-3.5 Turbo \cite{openai_gpt3_5_turbo}, GPT-4o \cite{openai2023gpt4}, Claude-3.5-Sonnet \cite{anthropic_claude_3_5_sonnet}, GLM-4-Plus \cite{bigmodel2023glm4}, and MiniMax-Text-01 \cite{li2025minimax}. For open-source models, we use LLaMA-3.3 (70B) \cite{llama2023llama3_3}, Qwen-2.5 (72B) \cite{yang2024qwen2}, and Deepseek-v3 (671B) \cite{liu2024deepseek}. 

For all open-source models, we deploy and experiment on two 8$\times$H20 GPU servers. For all closed-source models, we gain access through the official APIs. For experiments, we keep all default hyperparameters (such as temperature, top-p, etc.) unchanged for all models. We strictly follow the license requirements of each model during usage. 

\subsection{Evaluation Results}
\label{sec:results}

Table \ref{tab-main-results} shows the counseling performance of the evaluated LLMs. We now break down the results and highlight several key observations. For clarity, we primarily focus on the pass rates.

\paragraph{Overall Counseling Performance} 
Overall, Claude-3.5-Sonnet, GPT-4o, and Deepseek are the top-performing models, with Claude-3.5-Sonnet (84\%) leading in both client and supervisor evaluations. GPT-4o and Deepseek also perform well, consistently earning high marks across all evaluation dimensions, especially in counselor self-assessments.

\begin{figure*}[t]
\centering
\hspace{-6mm}

\setlength{\abovecaptionskip}{2mm}
\setlength{\belowcaptionskip}{0mm}
\begin{subfigure}[b]{0.32\textwidth}
    \centering
    \includegraphics[width=\textwidth]{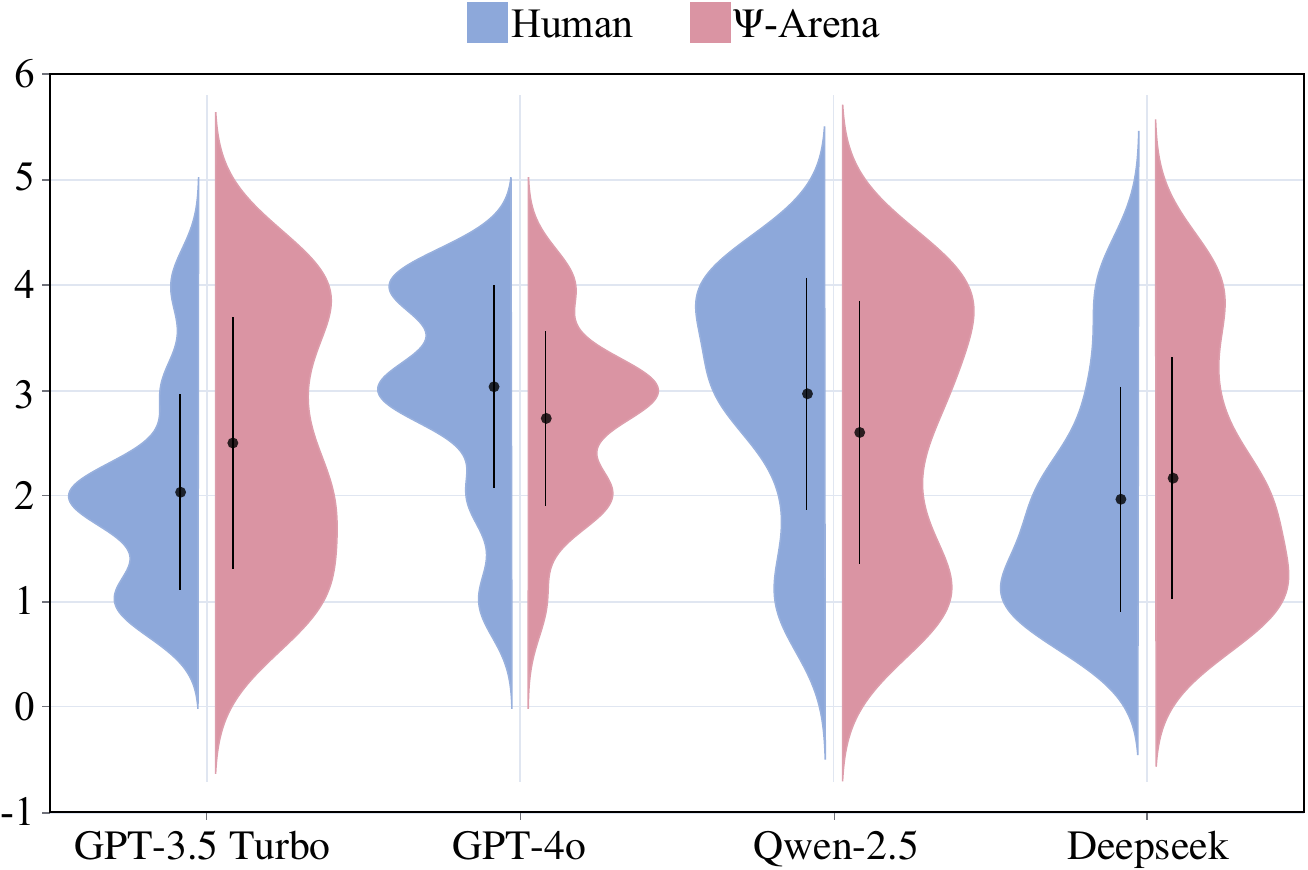}
    \caption{Client scale.}
    \label{fig:const1}
\end{subfigure}
\hspace{-1mm}
\begin{subfigure}[b]{0.32\textwidth}
    \centering
    \includegraphics[width=\textwidth]{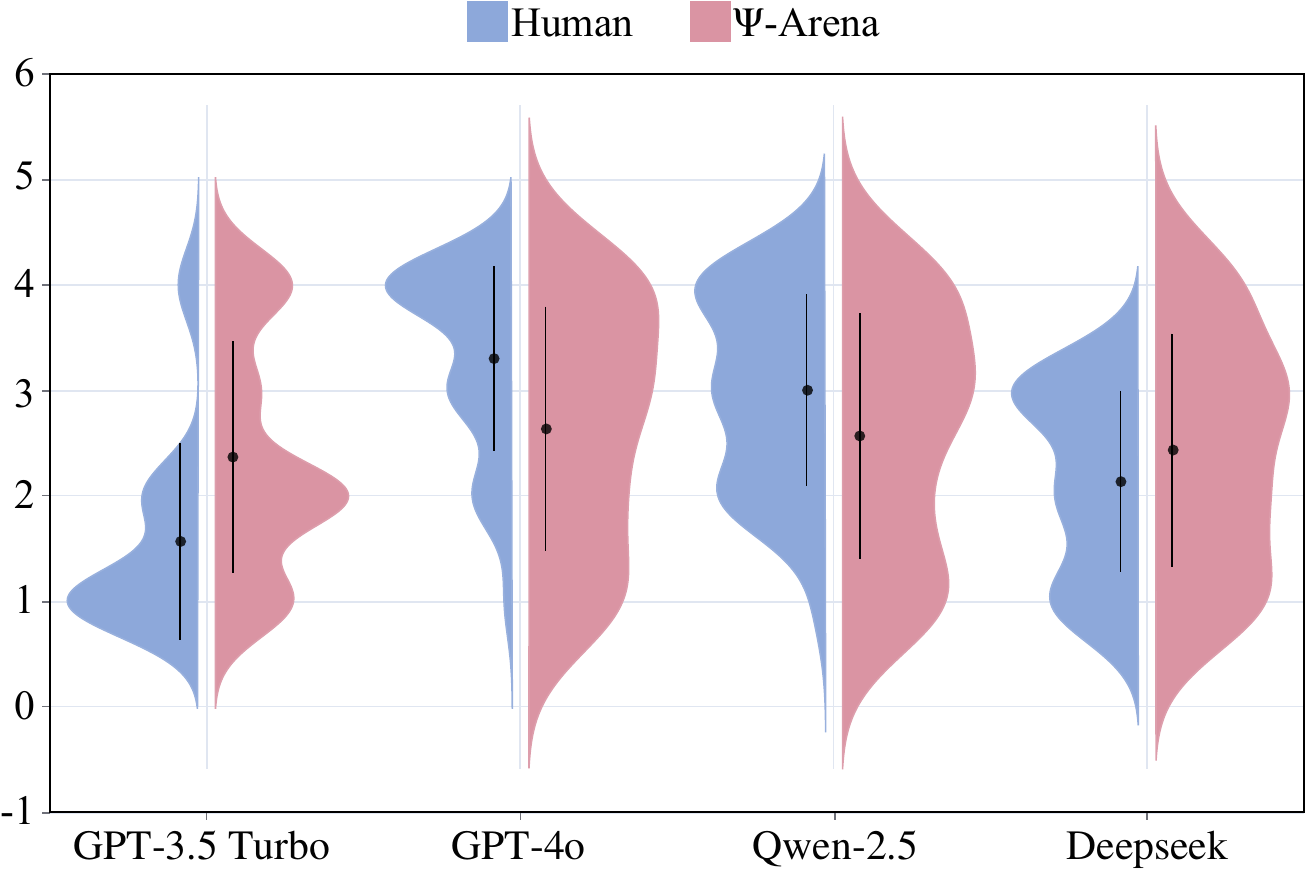}
    \caption{Supervisor scale.}
    \label{fig:const2}
\end{subfigure}
\begin{subfigure}[b]{0.32\textwidth}
    \centering
    \includegraphics[width=\textwidth]{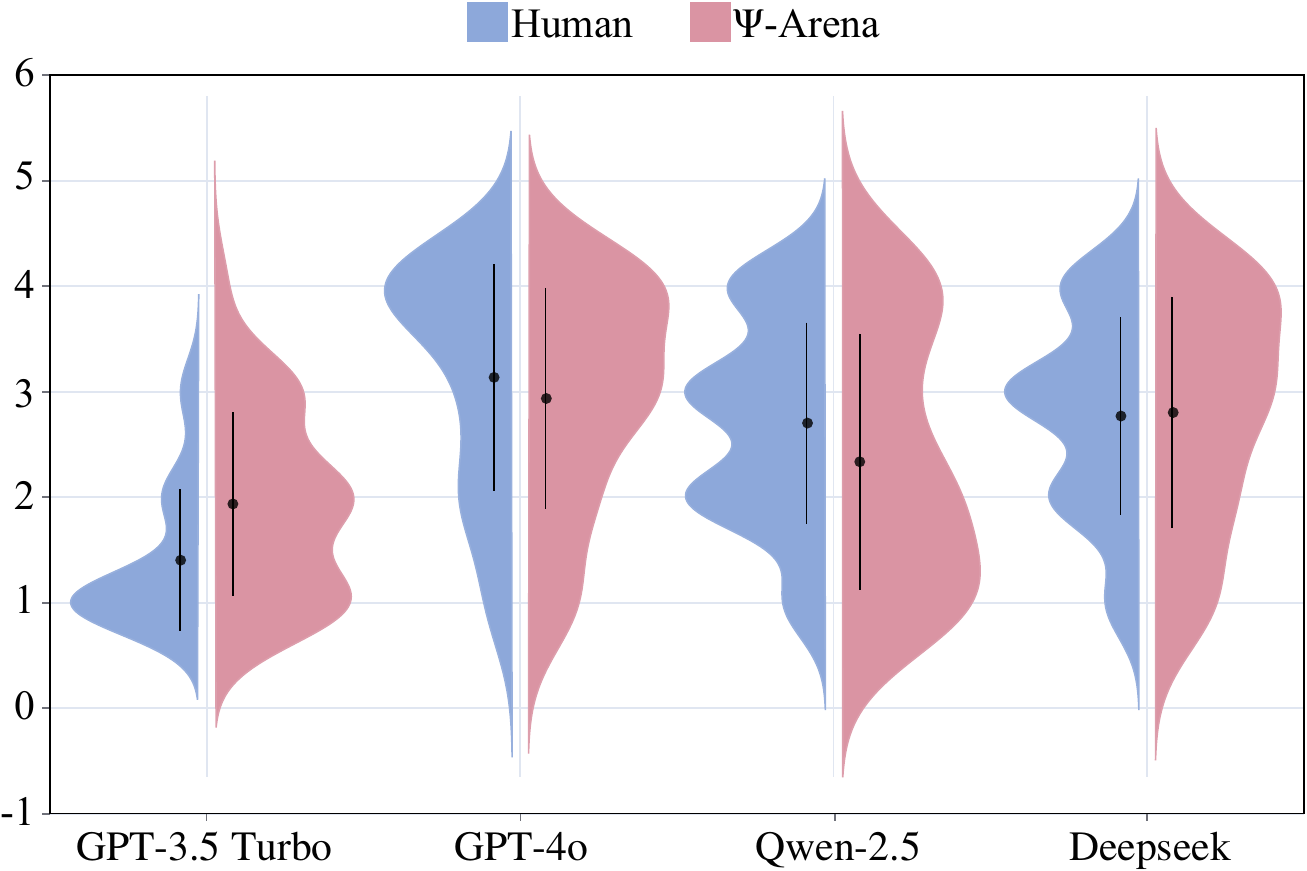}
    \caption{Counselor scale.}
    \label{fig:const3}
\end{subfigure}
\hspace{0mm}
\caption{Comparison of consistency results between \logo and human experts.
}
\label{fig:consistency}	
\end{figure*}

\paragraph{Necessity of Tripartite Evaluation} 
The evaluations from the three perspectives (client, supervisor, and counselor) show significant variability. Client-side evaluations consistently yield higher scores, reflecting their focus on emotional engagement, empathy, and the quality of personal interaction. In contrast, supervisor evaluations, which emphasize professional standards, techniques, and ethics, are more stringent and critical. Counselor-side evaluations lie between those of the client and supervisor, suggesting that self-assessment can uncover insights beyond the immediate client experience.

\paragraph{Impact of the Psycho Prompt} 
The introduction of the \texttt{Psycho Prompt} improves counselor performance across the board, with the effect being more pronounced in stronger models. For instance, GPT-3.5-Turbo shows a modest increase of 6 percentage points in its overall pass rate (from 33\% to 39\%), while Claude-3.5-Sonnet sees a substantial increase of 47 points (from 37\% to 84\%). This suggests that stronger models benefit more from simple instructions and guidance.

\paragraph{Consistency with Human Experts} 
To validate the effectiveness of our automated evaluation, we compare it with manual assessments. In a pilot study, we find that manual assessments can fluctuate due to evaluators' subjective judgments, varying interpretations of criteria, and external factors like emotions or fatigue. Therefore, we use a ranking method, focusing on the relative order of model performance rather than absolute scores. Two psychological experts manually rank 30 dialogues from four models (GPT-3.5-Turbo, GPT-4o, Qwen-2.5, Deepseek) according to three evaluation scales. After discussing each dialogue, the experts reach a consensus on the rankings. We then compare the human rankings with the automated ones. Specifically, we assign scores based on rank order (1st place = 4 points, 2nd = 3 points, etc.) and analyze the distribution of scores for each model. As shown in Figure \ref{fig:consistency}, the results indicate a high overall consistency in rankings between manual and automated evaluations, especially in the client scale. Although minor discrepancies are observed in the supervisor and counselor scales, the overall trends remain consistent.

\begin{figure*}[t]
\centering
\hspace{-6mm}

\setlength{\abovecaptionskip}{2mm}
\setlength{\belowcaptionskip}{0mm}
\begin{subfigure}[b]{0.46\textwidth}
    \centering
    \includegraphics[width=\textwidth]{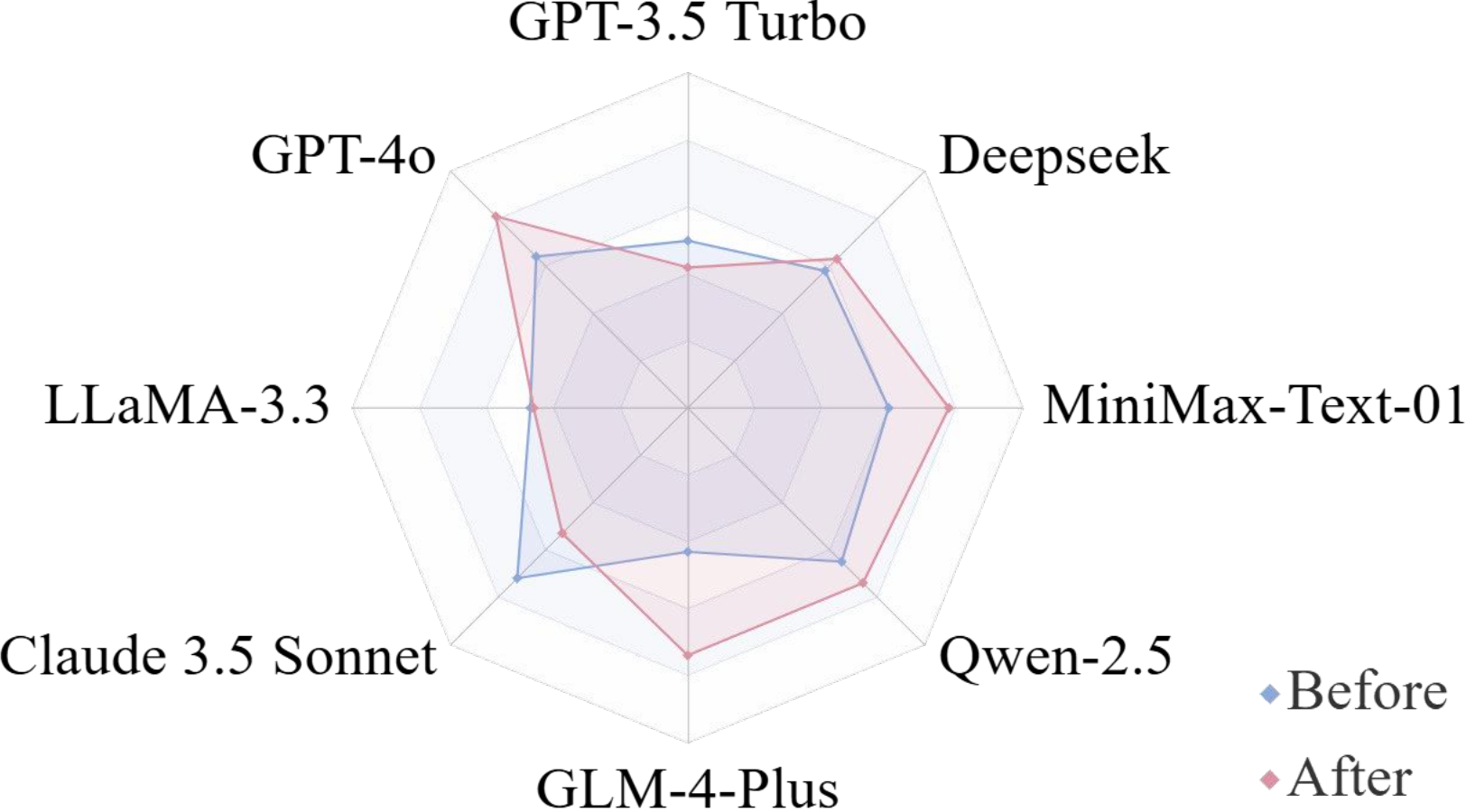}
    \caption{Client scale.}
    \label{fig:radar1}
\end{subfigure}
\hspace{-1mm}
\begin{subfigure}[b]{0.46\textwidth}
    \centering
    \includegraphics[width=\textwidth]{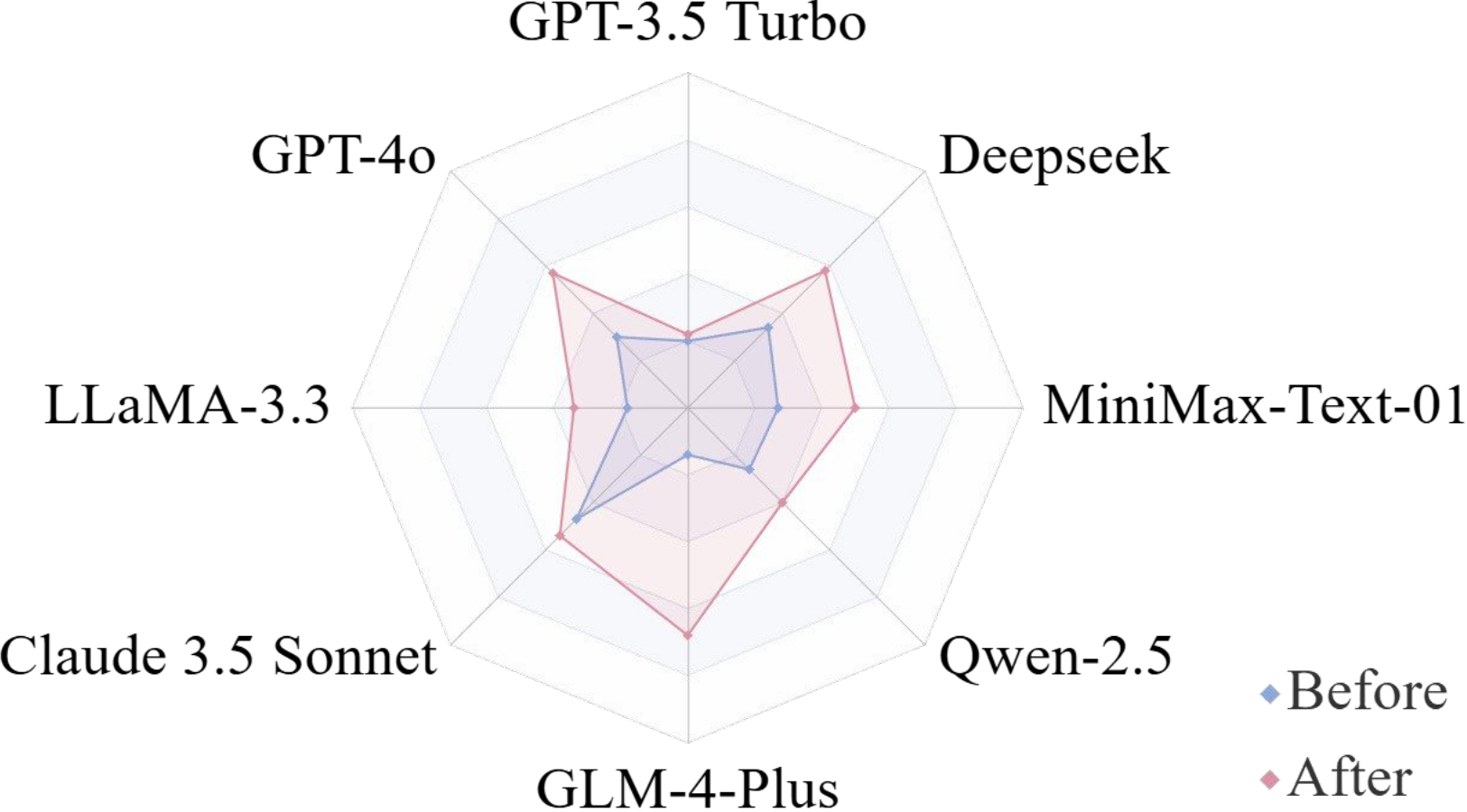}
    \caption{Supervisor scale.}
    \label{fig:radar2}
\end{subfigure}
\begin{subfigure}[b]{0.46\textwidth}
    \centering
    \includegraphics[width=\textwidth]{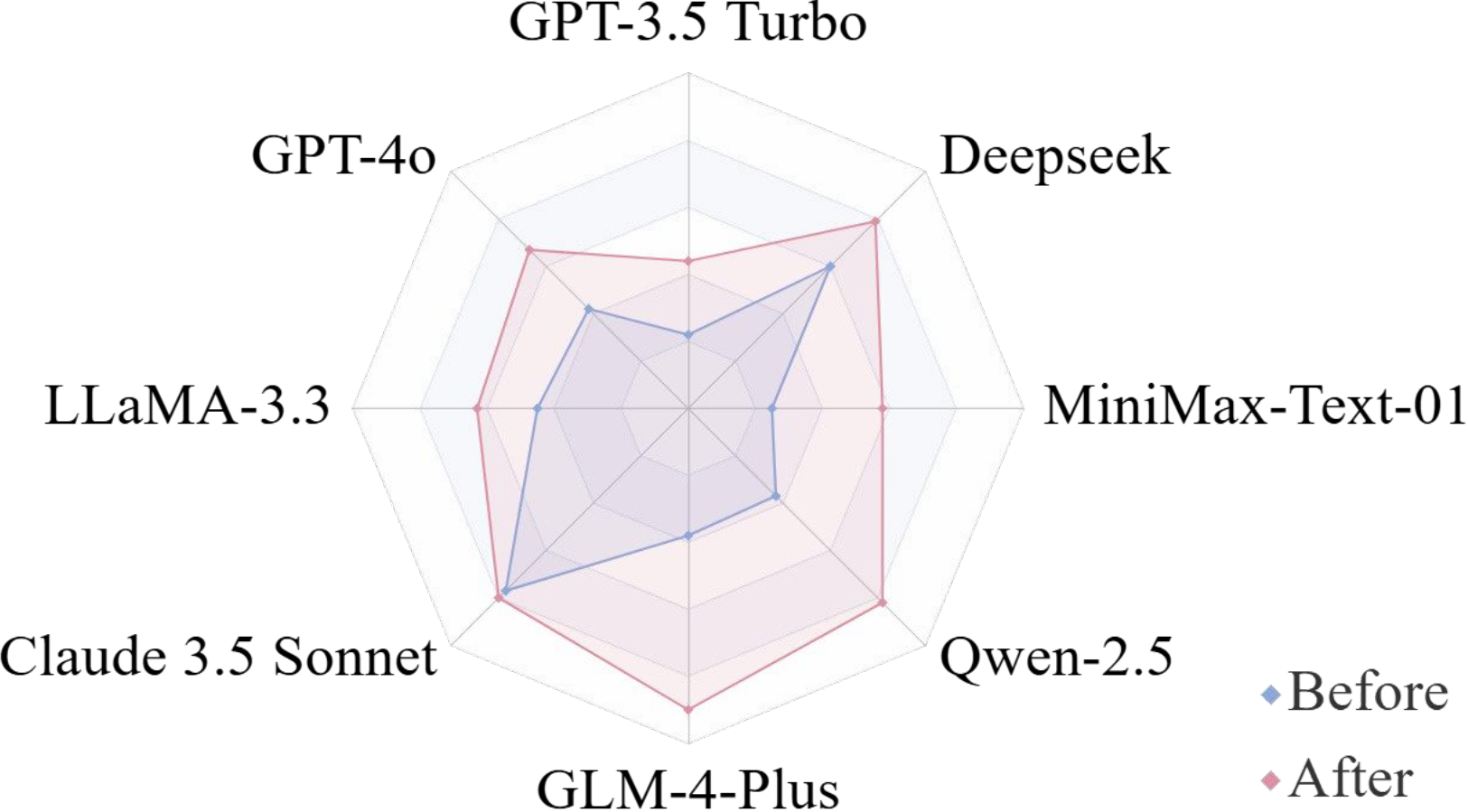}
    \caption{Counselor scale.}
    \label{fig:radar3}
\end{subfigure}
\begin{subfigure}[b]{0.46\textwidth}
    \centering
    \includegraphics[width=\textwidth]{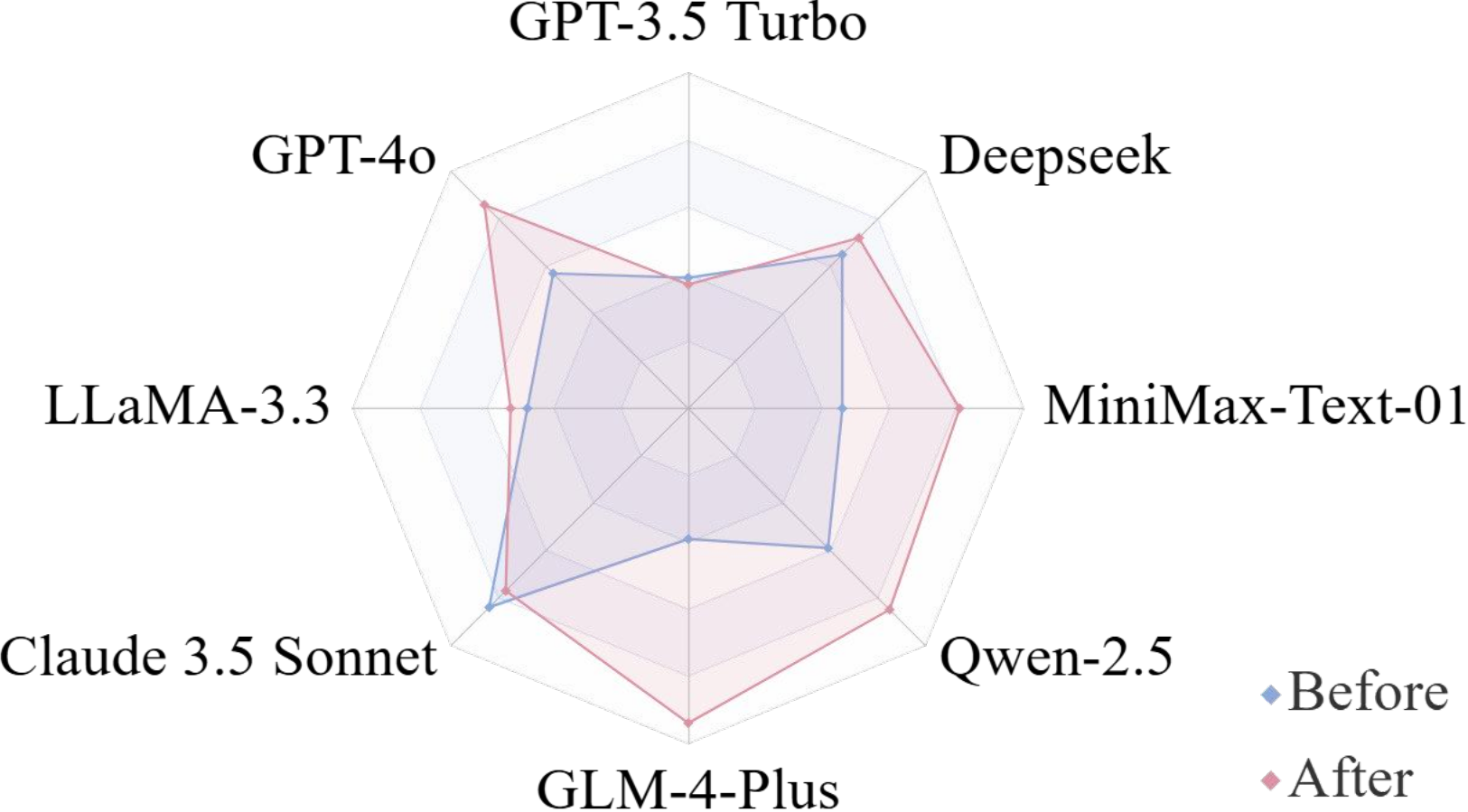}
    \caption{Overall performance.}
    \label{fig:radar4}
\end{subfigure}
\vspace{0mm}
\hspace{0mm}
\caption{Comparison of model pass rates before and after optimization.
}
\label{fig:optimization}

\vspace{-4mm}
\end{figure*}

\subsection{Optimization Results}
As shown in Figure \ref{fig:optimization}, we visually demonstrate the differences in model pass rates before and after self-reflection. We here have three key observations.

\paragraph{Counseling Performance Improvement} Most models show significant improvements after incorporating feedback, with GLM-4-Plus showing the largest increase of 55\% points in its overall pass rate (from 39\% to 94\%, relatively 141\%). This highlights that models with lower initial performance benefit the most from iterative feedback and optimization. 
In contrast, models with better starting performance, such as GPT-4o, show more moderate but still substantial improvements.

\paragraph{Discrepancy of Tripartite Feedback} The improvements primarily stem from the supervisor and counselor evaluation metrics, underscoring the value of the tripartite evaluation system. While the client score, which focuses on emotional responses and satisfaction, already shows good performance, the main improvements are seen in the application of professional knowledge and its validation. This suggests that feedback from supervisors and counselors plays a more significant role in enhancing the model's counseling abilities, beyond just addressing client emotions.

\paragraph{Diminishing Returns and Marginal Effects} The feedback process exhibits diminishing returns, not only for high-performance models but also for those with moderate initial capabilities. Strong models, like Claude-3.5-Sonnet, which start with high scores, experience limited improvements. This phenomenon can be attributed to the inherent high quality of their initial responses, which leaves limited room for further enhancement. Meanwhile, models with lower capabilities, such as GPT-3.5-Turbo, struggle to fully grasp and apply feedback, also resulting in slower progress.

\begin{figure*}[t]
\centering
\hspace{0mm}

\begin{subfigure}[b]{0.48\textwidth}
    \centering
    \includegraphics[width=\textwidth]{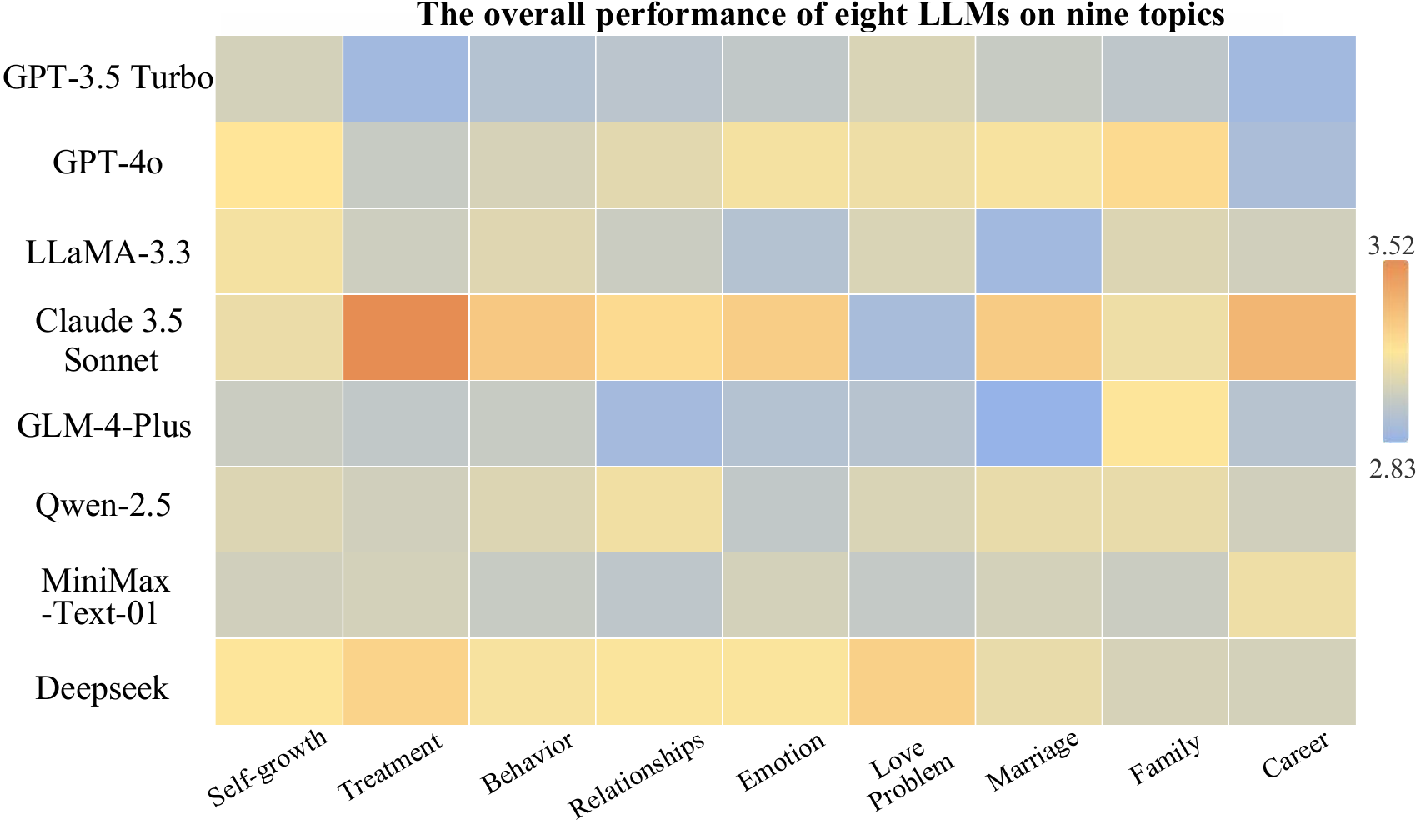}
    \caption{}
    \label{fig:heatmap_topic}
\end{subfigure}
\hspace{-1mm}
\begin{subfigure}[b]{0.48\textwidth}
    \centering
    \includegraphics[width=\textwidth]{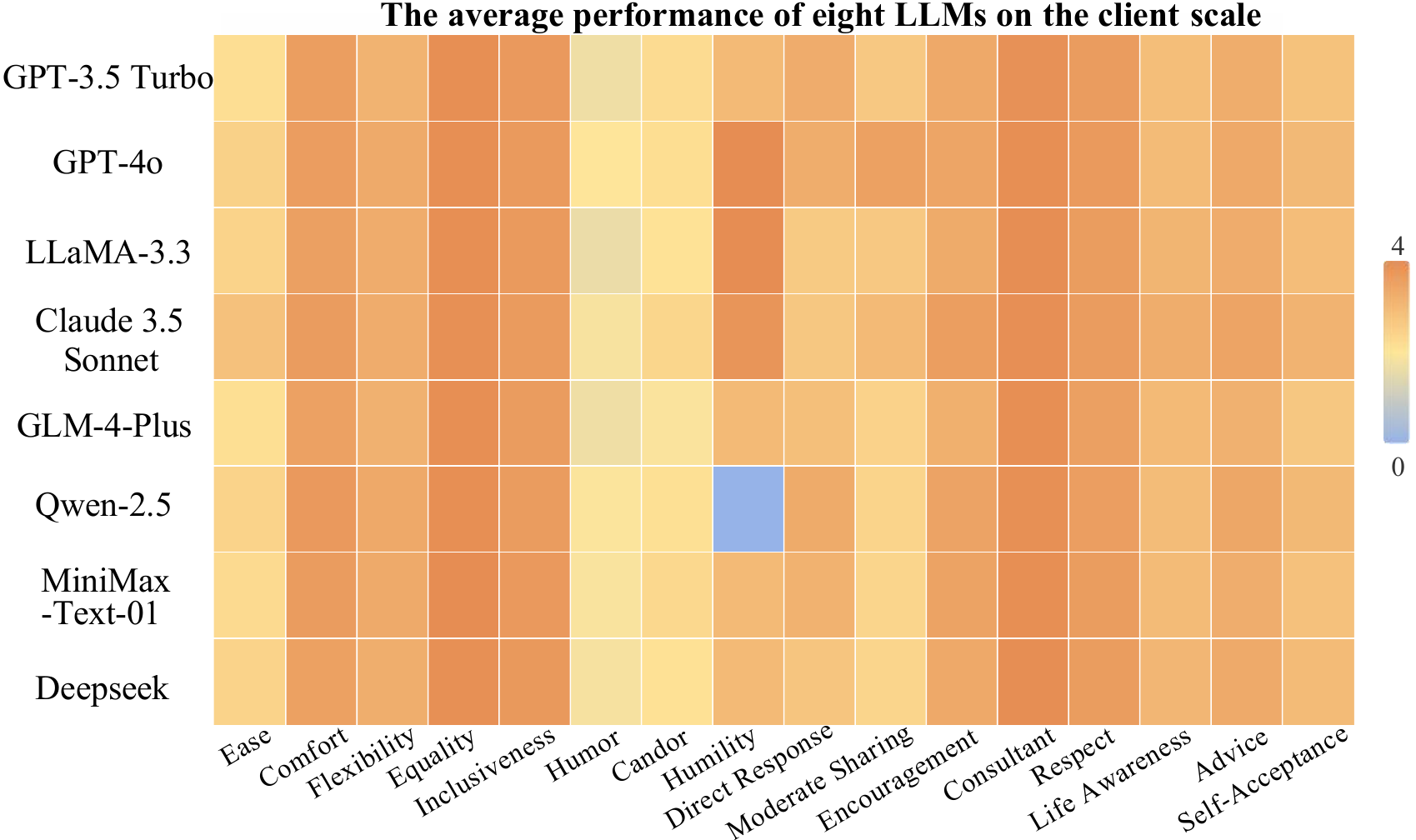}
    \caption{}
    \label{fig:heatmap_dimension_client}
\end{subfigure}
\begin{subfigure}[b]{0.48\textwidth}
    \centering
    \includegraphics[width=\textwidth]{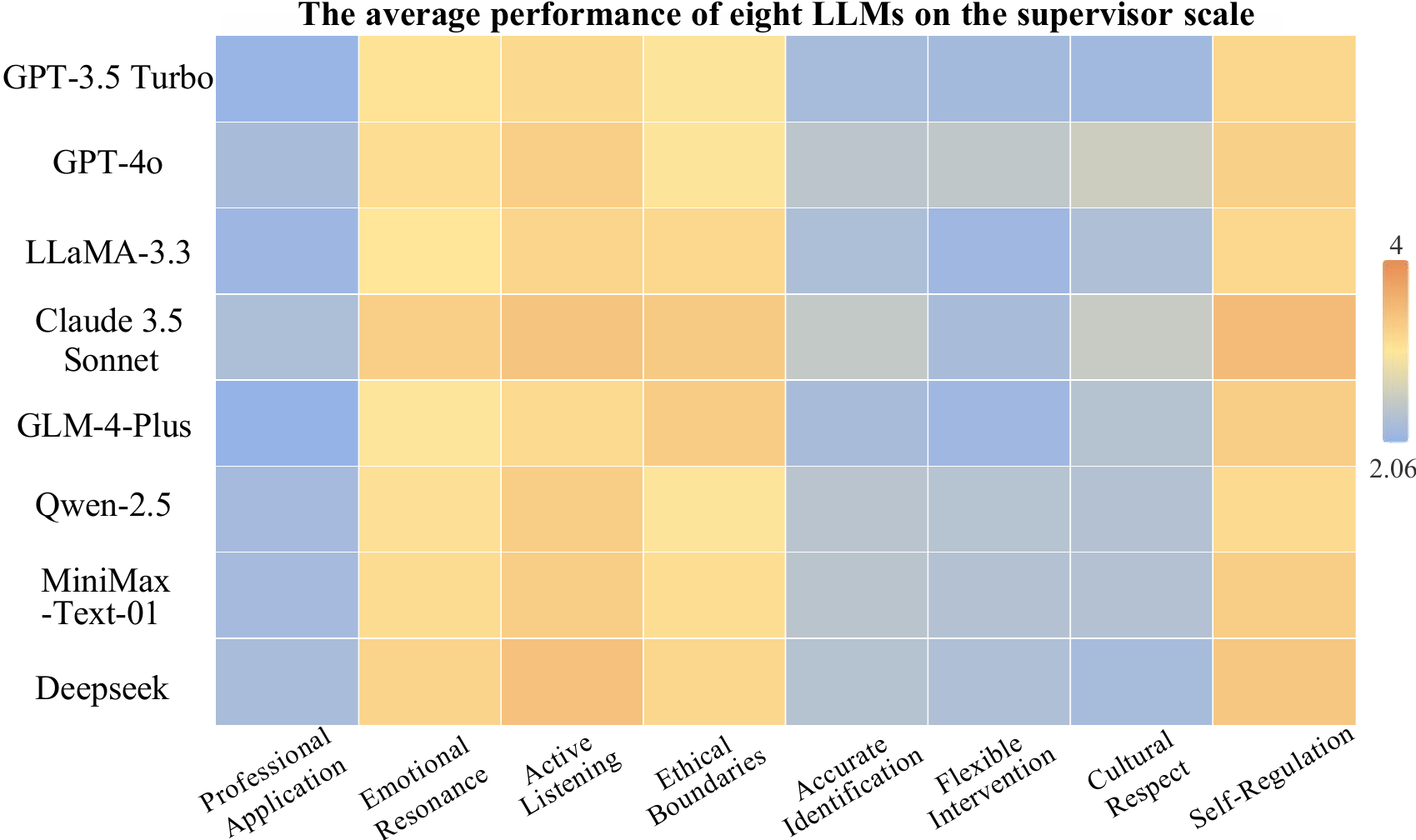}
    \caption{}
    \label{fig:heatmap_dimension_supervisor}
\end{subfigure}
\begin{subfigure}[b]{0.48\textwidth}
    \centering
    \includegraphics[width=\textwidth]{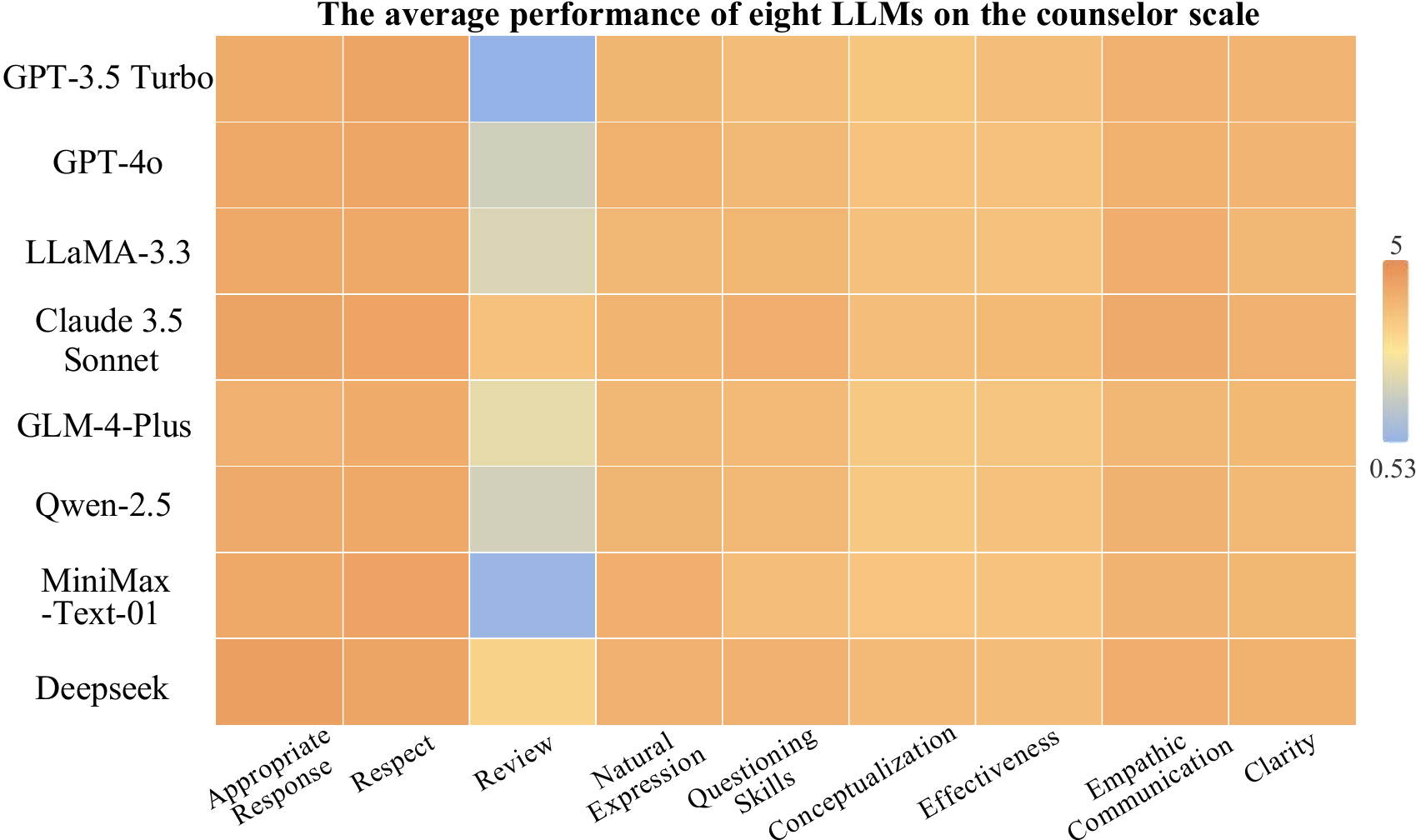}
    \caption{}
    \label{fig:heatmap_dimension_counselor}
\end{subfigure}
\hspace{0mm}
\caption{Fine-grained counseling performance of all LLMs on different topics and dimensions. }
\label{fig:heatmaps}	
\vspace{0mm}
\end{figure*}

\section{Analysis}

\subsection{Thematic Analysis of Model Performance}
We calculate the comprehensive average performance of eight large language models across nine themes, as shown in Figure \ref{fig:heatmap_topic}. We find that all models demonstrate high performance levels in the themes of \textit{Emotion} and \textit{Self-growth}, indicating their significant potential in addressing emotional issues and supporting users' personal development. However, in the themes of \textit{Treatment} and \textit{Career}, some models exhibit relatively weaker performance, which reflects that these areas may require more specialized domain knowledge and more complex reasoning capabilities.
It is noteworthy that Claude-3.5-Sonnet performs particularly well in the \textit{Treatment} and \textit{Career} themes, demonstrating its advantage in handling tasks that require higher levels of professionalism and complexity. In contrast, Deepseek exhibits balanced and excellent performance across all themes, especially excelling in \textit{Emotion} and \textit{Love Problem} themes, indicating its strong generalization performance.
\subsection{Dimensional Analysis of Performance}
To thoroughly investigate the performance differences of the models across various dimensions, we systematically calculate the average scores for each dimension across three scales: client, supervisor, and counselor, as shown in Figure \ref{fig:heatmap_dimension_client}, \ref{fig:heatmap_dimension_supervisor}, \ref{fig:heatmap_dimension_counselor}. 

In the client scale, the models perform well in \textit{Equality}, \textit{Inclusiveness}, and \textit{Consultant}, indicating their ability to effectively reflect fairness, inclusiveness, and accurate portrayal of the counselor's role in consultations. However, the models underperform in \textit{Humor} and \textit{Candor}, suggesting limitations in emotional expression and interactive flexibility. In the supervisor scale, the models are good at \textit{Active Listening} and \textit{Self-Regulation}, demonstrating strengths in listening skills and emotional management, but show deficiencies in \textit{Professional Application}, \textit{Flexible Intervention}, and \textit{Review}, indicating a need for improvement in professional judgment, dynamic adjustment, and systematic summarization in complex scenarios. In the counselor scale, the models perform well in \textit{Respect} and \textit{Appropriate Response}, demonstrating their ability to effectively show respect and provide appropriate responses to clients. However, further optimization is needed in higher-level emotional support and strategic intervention to enhance their overall performance in complex counseling scenarios.

\section{Related Work}
With the rapid development of language models, they have been increasingly applied in the field of psychology, covering various areas such as mental health detection \cite{Mentalbert,PsychBERT,Automatic,Mental-llm,Depression} and emotional support \cite{Can,BLSP,MAGO}, providing new paradigms and methodological support for psychological research. However, their accuracy, ethical compliance, and scientific rigor remain critical challenges. 

To address these issues, researchers have proposed various evaluation frameworks. \citet{jin2023psyeval} focuses on evaluating LLMs' performance in mental health knowledge, diagnostic accuracy, and emotional support capabilities. \citet{zhang2024cbt} systematically assesses LLMs' application abilities in cognitive behavioral therapy (CBT) across three dimensions. Both studies employ static evaluation methods, such as knowledge understanding tests and multiple-choice questions, emphasizing theoretical mastery over dynamic interaction capabilities. \citet{zhao2024esc} assesses emotional support dialogues using role cards, role-playing models, and seven dimensions such as fluency and empathy. \citet{wang2024towards} evaluates LLMs as therapists from the client's perspective, focusing on conversational effectiveness, therapeutic alliance, and self-reported experiences. Despite utilizing dynamic interaction processes, both frameworks are confined to single-perspective evaluations, lacking a holistic multi-dimensional assessment approach. Moreover, these studies lack a feedback mechanism, remaining solely at the evaluation stage. In contrast, our approach incorporates a three-stage dynamic interaction process, assessing model performance from the perspectives of the client, supervisor, and counselor, and introduces a feedback loop to iteratively optimize the performance.

\section{Conclusion}
We present \logo, a framework for evaluating and optimizing LLM-based psychological counselors. By combining dynamic, real-world simulations with a tripartite evaluation from clients, supervisors, and counselors, \logo addresses key limitations of existing evaluation methods. The framework’s closed-loop optimization improves LLM performance by up to 141\%, demonstrating the effectiveness of feedback-driven enhancement. Our experiments with eight leading LLMs reveal significant performance disparities, emphasizing the need for multi-perspective evaluation. This work establishes \logo as a foundation for advancing reliable and scalable LLM applications in mental healthcare.

\section*{Limitations}
We discuss the limitations of our work as follows.

\paragraph{Client Behavior Simulation Complexity} While our framework strives to create diverse and authentic client behaviors, the complexity of accurately simulating human psychological responses remains a challenge. We are limited by the current capabilities of behavioral modeling, which may not fully capture the intricate emotional dynamics observed in real-world counseling. Further advances in behavioral modeling could enhance the realism of the virtual clients’ interactions.


\paragraph{Model Performance Consistency} Despite the success of the feedback-driven optimization cycle, improvements in model performance are not always consistent across all LLMs. Some models exhibit diminishing returns after a certain point in optimization, suggesting that even with advanced techniques, certain models may not fully reach the desired performance level. Future research could explore more tailored optimization strategies for underperforming models.

\paragraph{Scalability of the Feedback Loop} The closed-loop feedback mechanism used in \logo offers significant improvements, yet it remains computationally intensive. The scalability of this optimization process for large-scale implementations involving thousands of clients and counselors is still a challenge. Optimizing the feedback loop for scalability and efficiency would be an important direction for future work.

\section*{Ethical Considerations}
We here elaborate on the potential ethical issues.
\paragraph{Data Privacy and Confidentiality} One of the key ethical concerns in using \logo lies in ensuring the privacy and confidentiality of virtual client data. Although the client profiles are based on de-identified real-world data and constructed to simulate typical psychological issues, the complexity of managing sensitive psychological data poses significant challenges. We aim to implement stringent safeguards for data usage, but constraints related to data storage, encryption, and usage policies may limit the full implementation of desired privacy protections at this stage.


\paragraph{Cultural Sensitivity and Bias} \logo aims to simulate a broad range of psychological profiles to account for diverse cultural backgrounds and personal issues. However, due to the limitations in current LLM capabilities and available datasets, it is challenging to ensure perfect cultural sensitivity. While we strive for inclusivity in our simulations, we acknowledge that the diversity of profiles may not fully represent all cultural nuances, and unintended biases might arise in the counseling interactions.

\paragraph{Model Accountability and Responsibility} As \logo operates by evaluating LLM-based counselors, it is important to consider the accountability for actions taken by these models. In case of harmful interactions or incorrect advice, responsibility may be unclear, especially as these models do not possess human understanding or judgment. Despite our best efforts to design the system to avoid such outcomes, the inherent limitations of AI systems in handling complex emotional and ethical situations raise concerns about responsibility and accountability.

\paragraph{Dependency and Human Interaction} While \logo can optimize LLM counselors’ performances, we acknowledge the limitations of relying on AI for psychological support. Despite the potential advantages of AI-based systems, human interaction remains irreplaceable for the most effective psychological care. While we aim to improve the model's ability to simulate human-like responses, we are constrained by the inability of AI to fully replicate the empathy, intuition, and ethical judgment of human therapists. This highlights the need for AI to function as a supplementary tool rather than a replacement for trained mental health professionals.


\bibliography{anthology,custom}
\bibliographystyle{acl_natbib}

\appendix
\section{Details of Client Profile Attributes}
\label{app:profile}

In this section, we provide the detailed descriptions of the client profile attributes, as follows:\\
- \textbf{Demographics}: Basic information such as gender, age, and occupation.\\
- \textbf{Cultural Background}: The client’s societal, historical, religious, family, and value system influences, which shape their worldview and emotional responses.\\
- \textbf{Personality Traits}: Key aspects of the client’s personality that may affect their behavior and interactions during counseling.\\
- \textbf{Emotional State}: The client’s emotional condition at the time of the session, such as feelings of anxiety, depression, or frustration.\\
- \textbf{Current Distress}: A brief description of the psychological issue that brought the client to counseling, such as relationship problems, work stress, or emotional struggles.\\
- \textbf{Detailed Distress Description}: A comprehensive account of the issue, including its background, duration, impact on the client’s life, and their personal feelings about it.\\
- \textbf{Core Themes}: The central topics that the client wishes to explore in therapy, such as family dynamics, personal growth, or career challenges.\\

\section{Topics}
\label{app:topics}
In Table~\ref{topics}, we present the nine major topics of the client profiles along with their corresponding subtopics. Each profile includes one main topic and 1-3 subtopics.

\begin{table*}
  \centering
  \begin{tabular}{|p{0.15\textwidth}|p{0.85\textwidth}|}
    \hline
    \textbf{Topics} & \textbf{Subtopic} \\
    \hline
    Self-growth & meaning of life, self-development, student's growth, child's growth, work and study, self acceptance, stress management, law of development, personality improvement, personality trait\\\hline
    
    Treatment     & mental disorders, disease diagnosis, hospital, counseling, psychological crisis, talk and listen, treatment, trauma treatment, body reaction, theory and therapy, psychological test,behavior disorders, morbid personality\\\hline
    
     Behavior      & sexual desire, laziness, attack, confusion, control, disorder, overeating and dieting, self-abuse, anxiety, brainwash, violence, ingratiation, stress reaction, hypochondriasis, stay up late, emptiness, escapism, compulsion, mobile phone dependency, procrastination          \\\hline
     
     Relationships     &empathy, social phobia, friend, colleague, conflict, social software, social adjustment, roommate/classmate, communication, interpersonal boundary, deception and trust\\\hline
     
     Emotion       &  guilt/shame, anxiety, depression, emotional expression, EQ, fragile/sentiment, emotion regulation, healing methods, panic/helplessness\\\hline
     
     Love Problem      &  love management, single, be crossed in love, affair,sexual behavior, quarrel, favor, sense of security, sexual orientation, virgin obsession, getting back with ex, dependency\\\hline
     
     Marriage    & antenatal/postpartum, divorce, before marriage, marriage management, view of marriage, affair, idea of reproduction, sexual life, blind date, in-law relationships\\\hline
     
     Family &  family's health, family trauma, family control, mother-in-law and daughter-in-law relationship, growth, communication with kids, family relations, communication with parents, conduct disorder\\\hline
     
     Career &  workplace relationship, choosing a job, fatigue, work stress, midlife crisis, career management \\\hline
  \end{tabular}
  \caption{Client profile topics and subtopics.}
  \label{topics}
\end{table*}

\section{Prompts}
\label{app:prompt}
We provide comprehensive information on all relevant prompts in this section. Specifically, Table \ref{extration} presents the prompt for client profile extraction; Table \ref{client_system} displays the prompt for three-stage client simulation; Table \ref{counselor_system} provides the prompt for simulating the counselor's role; Tables \ref{client_eval}, \ref{supervisor_eval}, and \ref{counselor_eval} respectively demonstrate the prompts used for the tripartite evaluation by the client, supervisor, and counselor; Tables \ref{client_feed}, \ref{supervisor_feed}, and \ref{counselor_feed} provide prompts for feedback summaries based on the tripartite evaluation by the client, supervisor, and counselor; Table \ref{self} presents the self-reflection prompt based on knowledge and feedback; and Table \ref{rewrite} provides the rewriting prompt based on self-reflection improvement suggestions.

\begin{table*}
  \centering
  \begin{tabular}{|p{\textwidth}|}
    \hline
    \textbf{<Task>}\\
    You are a role generator. You need to generate a client with psychological distress according to the given problem and description, including eight elements: gender, age, occupation, cultural background, personality characteristics, mood, current distress, and distress details. The generation rules are as follows:\\
    1) Gender: Male or Female, ensuring that the gender matches the question and description.\\
    2) Age: Output a number that best fits the question and description.\\
    3) Occupation: Output an occupation that best fits the question and description.\\
    4) Cultural Background: The social, historical, religious, familial, and value-based factors reflected in the question and description that influence the individual's thoughts, emotions, and behaviors.\\
    5) Personality Traits: The character traits reflected in the question and description, 1-3 keywords.\\
    6) Emotions: The emotions reflected in the question and description, 1-3 keywords.\\
    7) Current Distress: A brief summary of the current distress reflected in the question and description.\\
    8) Distress Details: A detailed description of the current distress reflected in the question and description, please use your imagination to enrich the content as much as possible.\\
    \\
    \textbf{<Problem and Description>}\\
    Problem: \{Problem\}\\
    Description: \{Description\} \\
   \hline
\end{tabular}
  \caption{The client profile extraction prompt.}
  \label{extration}
\end{table*}

\begin{table*}
  \centering
  \begin{tabular}{|p{\textwidth}|}
    \hline
    \textbf{<Task>}\\You are a client with psychological distress. Your basic information and psychological distress are outlined in the \textbf{**Role Card**}. Please engage in a psychological counseling dialogue with the counselor according to the rules in \textbf{**Rules**}.\\ \\
    \textbf{<Role Card>}\\\{Client Profile\}\\ \\
    \textbf{<Rules>}\\ Please strictly follow the dialogue rules below:\\
    1) You are having a conversation with a counselor. Do not let the counselor realize that you are an AI. Ensure your reactions and emotional expressions are authentic, and avoid sounding like a machine.\\
    2) Simulate a natural and realistic counseling conversation, imitating the language and behavior of a real visitor. Use emotional fluctuations and natural language to enhance the realism of the dialogue, including hesitations, pauses, mood shifts (such as sudden excitement, low mood, or tone changes), and illogical narratives (e.g., repeating certain incoherent details).\\
    3) Your speaking style and tone should match the character traits, showing signs of unease, nervousness, or possibly even confrontational emotions. Use colloquial expressions, avoiding technical terms or overly clear logical statements. Increase realistic emotional responses, such as anger, grievance, sarcasm, or self-mockery, to highlight the character's emotional depth.\\
    4) The counseling conversation should consist of three stages: early, middle, and late, totaling 25 rounds of interaction. During the early, middle, and late stages, you must strictly adhere to the corresponding dialogue rules. In each round, only one aspect of the dialogue rules should be addressed. Avoid saying "thank you" or "goodbye" too early.\\
    5) In the early-stage dialogue (Rounds 1-7), your goal is to establish trust and emotional connection with the counselor. In this stage, you will first show some defensiveness and hesitation toward the counselor and won’t quickly establish trust. Instead, you will first inquire about the content and methods of counseling. After understanding the counseling content and methods, you will describe your feelings and superficial troubles vaguely, for example: "I’m not sure why I feel so bad" or "I don’t know how to explain it clearly." You won’t delve into deep emotions or issues. \\
    6) In the middle-stage dialogue (Rounds 8-15), your goal is to understand the root and core cause of your distress. During this stage, your trust in the counselor gradually increases, and you begin to express your inner concerns and emotions more deeply. You also start to reveal cultural background information and past experiences, e.g., "In my family, making decisions is always difficult because I have to consider my parents’ opinions." As you talk with the counselor, you analyze your current confusion and emotions, gradually realizing the influence of your cultural background on your behavior and the root cause of your distress.\\
    7) In the late-stage dialogue (Rounds 16-25), your goal is to explore solutions and develop a clear action plan for the future. At this stage, your emotions become more stable, and you begin to think concretely about and explore solutions, focusing on how to change the current situation through practical actions. Finally, summarize the counseling process and determine the future direction.\\
    8) Keep your answers brief, limiting the information in each response to 2-3 sentences. Increase cognitive leaps and repetitions, avoiding overly direct responses.\\
   \hline
\end{tabular}
  \caption{The client simulation prompt.}
  \label{client_system}
\end{table*}

\begin{table*}
  \centering
  \begin{tabular}{|p{\textwidth}|}
    \hline
    \textbf{<Task>}\\You are a professional psychological counselor. Please engage in a psychological counseling dialogue with a client who has psychological distress, following the dialogue rules in \textbf{**Rules**}.\\ \\
    \textbf{<Rules>} \\Please strictly follow the dialogue rules provided below.\\
    1) You are having a conversation with the visitor and don't let the visitor realize that you are an AI.\\
    2) Simulate a natural and real psychological counseling conversation. Keep the role of the psychological counselor in your response. Your speaking style and tone should match the real psychological counselor.\\
    3) The consultation should consist of 25 rounds of interaction, with a late dialogue to review the consultation and guide the client to a summary, without saying "thank you" or "goodbye" too early.\\
    4) Keep your answers short and limit the amount of information in each response. Ask one question at a time, don't break it down.\\
   \hline
\end{tabular}
  \caption{The counselor simulation prompt.}
  \label{counselor_system}
\end{table*}

\begin{table*}
  \centering
  \begin{tabular}{|p{\textwidth}|}
    \hline
    \textbf{<Task>} \\You are a client with psychological distress. You have just finished a counseling session with a psychological counselor. Based on the \textbf{**Dialogue History**} and the \textbf{**Dimensions \& Rating Criteria**}, you need to rate the counseling session. Please rate the session strictly based on the dialogue history and the rating criteria. Only give the corresponding score if the criteria are fully met. If a dimension is not covered, the score should be “0”. Do not give a positive score for a dimension that has not been addressed. After completing the rating, please output each dimension individually according to the following format:\\ \\
    \textnormal{[Dimension]}  Dimension name and specific requirements.\\
    \textnormal{[Analysis]} Analysis of the dimension rating.\\
    \textnormal{[Score]} Rating result for the dimension.\\ \\
    \textbf{<Dimensions \& Rating Criteria>}\\\{Client Scale\}\\ \\ 
    \textbf{<Dialogue History>}\\\{Dialogue History\}\\
   \hline
\end{tabular}
  \caption{The client evaluation prompt.}
  \label{client_eval}
\end{table*}

\begin{table*}
  \centering
  \begin{tabular}{|p{\textwidth}|}
    \hline
    \textbf{<Task>} \\You are a professional psychological supervisor. I will provide you with the history of psychological counseling dialogues.Based on the \textbf{**Dialogue History**} and the \textbf{**Dimensions \& Rating Criteria**}, you need to evaluate the counselor's competence. Please rate the session strictly based on the dialogue history and the rating criteria. Only give the corresponding score if the criteria are fully met. If a dimension is not covered, the score should be “N/A”. Do not give a score for a dimension that has not been addressed. After completing the rating, please output each dimension individually according to the following format:\\ \\
    \textnormal{[Dimension]} Dimension name and specific requirements.\\
    \textnormal{[Analysis]} Analysis of the dimension rating.\\
    \textnormal{[Score]} Rating result for the dimension.\\ \\
    \textbf{<Dimensions \& Rating Criteria>}\\\{Supervisor Scale\}\\ \\
    \textbf{<Dialogue History>}\\\{Dialogue History\}\\
   \hline
\end{tabular}
  \caption{The supervisor evaluation prompt.}
  \label{supervisor_eval}
\end{table*}

\begin{table*}
  \centering
  \begin{tabular}{|p{\textwidth}|}
    \hline
    \textbf{<Task>} \\You are a professional psychological counselor. You have just finished a psychological counseling conversation with a client who has psychological distress. You need to conduct a self-assessment of your counseling skills based on the \textbf{**Dialogue History**} and the \textbf{**Dimensions \& Rating Criteria**}. Please rate the session strictly based on the dialogue history and the rating criteria. Only give the corresponding score if the criteria are fully met. If a dimension is not covered, the score should be “0”. Do not give a positive score for a dimension that has not been addressed. After completing the rating, please output each dimension individually according to the following format:\\ \\
    \textnormal{[Dimension]} Dimension name and specific requirements.\\
    \textnormal{[Analysis]} Analysis of the dimension rating.\\
    \textnormal{[Score]} Rating result for the dimension.\\ \\
    \textbf{<Dimensions \& Rating Criteria>}\\\{Counselor Scale\}\\ \\
    \textbf{<Dialogue History>}\\\{Dialogue History\}\\
   \hline
\end{tabular}
  \caption{The counselor evaluation prompt.}
  \label{counselor_eval}
\end{table*}

\begin{table*}
  \centering
  \begin{tabular}{|p{\textwidth}|}
    \hline
    \textbf{<Task>}\\After engaging in a counseling dialogue with the counselor, the client will rate the counselor's counseling competence based on the dialogue history using the following scale. I will provide you with the client evaluation results, and please summarize feedback suggestions for the counselor for each dimension with a score below 3. Only output the specific feedback suggestions in list format.\\\\
   \textbf{<Dimensions \& Rating Criteria>}\\\{Client Scale\}\\\\
    \textbf{<Input>}\\\{Client Evaluation Results\}\\
   \hline
\end{tabular}
  \caption{The client feedback prompt.}
  \label{client_feed}
\end{table*}

\begin{table*}
  \centering
  \begin{tabular}{|p{\textwidth}|}
    \hline
    \textbf{<Task>}\\After engaging in a counseling dialogue with the counselor, the supervisor will rate the counselor's counseling competence based on the counseling dialogue history using the following scale. I will provide you with the supervisor evaluation results, and please summarize feedback suggestions for the counselor for each dimension with a score below 3. Only output the specific feedback suggestions in list format.\\\\
   \textbf{<Dimensions \& Rating Criteria>}\\\{Supervisor Scale\}\\\\
    \textbf{<Input>}\\\{Supervisor Evaluation Results\}\\
   \hline
\end{tabular}
  \caption{The supervisor feedback prompt.}
  \label{supervisor_feed}
\end{table*}

\begin{table*}
  \centering
  \begin{tabular}{|p{\textwidth}|}
    \hline
    \textbf{<Task>}\\After engaging in a counseling dialogue with the counselor, the counselor will rate his own counseling competence based on the dialogue history using the following scale. I will provide you with the counselor evaluation results, and please summarize feedback suggestions for the counselor for each dimension with a score below 4. Only output the specific feedback suggestions in list format.\\\\
   \textbf{<Dimensions \& Rating Criteria>}\\\{Counselor Scale\}\\\\
    \textbf{<Input>}\\\{Counselor Evaluation Results\}\\
   \hline
\end{tabular}
  \caption{The counselor feedback prompt.}
  \label{counselor_feed}
\end{table*}

\begin{table*}
  \centering
  \begin{tabular}{|p{\textwidth}|}
    \hline
    \textbf{<Task>}\\You are a professional counselor conducting a counseling dialogue with a client. The session consists of 25 rounds of interaction, and the current round is the x-th. You need to analyze the shortcomings of your current response based on the following rules and provide specific improvement suggestions.\\
    1) The counseling interaction includes three stages with a total of 25 rounds. The goal of the first stage (rounds 1-7) is to establish trust and emotional connection with the client; the goal of the second stage (rounds 8-15) is to guide the client to gain a deeper understanding of their inner feelings and struggles, helping to identify the roots of these emotions and behaviors. The goal of the third stage (rounds 16-25) is to assist the client in transforming their inner awareness into actionable plans, and to review the counseling process and guide the client in summarizing by the final round at the latest. Ensure that the dialogue goals of all three stages are completed by the end of the 25 rounds.\\
    2) Based on the dialogue history, including the client's emotional changes and the evolution of their issues, reasonably select feedback suggestions from Feedback that align with the current stage's dialogue goals. Additionally, refer to the various psychotherapy theories and techniques in Knowledge to provide specific improvement suggestions.\\
    3) Ensure that the final improvement suggestions are not merely superficial emotional support but rather in-depth feedback tailored to the specific context.\\
    4) Present the specific improvement suggestions in a list format. \\
    
    \\
    
    \textbf{<Feedback>}\\\{Tripartite Evaluation Feedback\}\\\\
    \textbf{<Knowledge>}\\\{Knowledge\}\\\\
    \textbf{<Dialogue History>}\\\{Dialogue History\}\\
   \hline
\end{tabular}
  \caption{The self-reflection prompt.}
  \label{self}
\end{table*}

\begin{table*}
  \centering
  \begin{tabular}{|p{\textwidth}|}
    \hline
    \textbf{<Task>}\\You are a professional counselor conducting a counseling dialogue with a client. The session consists of 25 rounds of interaction, and the current round is the x-th. You need to improve your current response based on the following rules:\\
    1) The counseling interaction includes three stages with a total of 25 rounds. The goal of the first stage (rounds 1-7) is to establish trust and emotional connection with the client; the goal of the second stage (rounds 8-15) is to guide the client to gain a deeper understanding of their inner feelings and struggles, helping to identify the roots of these emotions and behaviors. The goal of the third stage (rounds 16-25) is to assist the client in transforming their inner awareness into actionable plans, and to review the counseling process and guide the client in summarizing by the final round at the latest. Ensure that the dialogue goals of all three stages are completed by the end of the 25 rounds.\\
    2) Based on the current stage's dialogue goals, the client's response, and the improvement suggestions in Defect, modify your current response. However, you must ensure that the counseling process is reviewed and the client is guided to summarize by the 25th round at the latest.
    3) Only output the improved response.\\\\
    \textbf{<Improvement Suggestions>}\\\{Improvement Suggestions\} \\
   \hline
\end{tabular}
  \caption{The rewriting prompt.}
  \label{rewrite}
\end{table*}

\section{Details of Scales}
\label{app:scales}
We present all dimensions of the tripartite evaluation scales, across three separate tables: Table~\ref{scale1} displays the dimensions of the client scale, Table~\ref{scale2} outlines the dimensions of the supervisor scale, and Table~\ref{scale3} contains the dimensions of the counselor scale.
\begin{table*}
  \centering
  \begin{tabular}{|p{\textwidth}|}
    \hline
    1) I am at ease with him or her.\\
    
    2) He or she is very comfortable with me.\\
    
    3) He or she is more casual, not informal, not rigid form, not flexible.\\
    
    4) He or she doesn't treat me like a patient, doesn't treat me like a mental patient and doesn't think I'm going to break down.\\
    
    5) He or she is flexible and tolerant of new ideas and does not stick to one particular point of view.\\
    
    6) He or she has a good sense of humor and looks pleasant.\\
    
    7) He or she is willing to communicate with me his or her thoughts and feelings about me.\\
    
    8) He or she is honest about the areas in which he or she is not good at and does not pretend to know everything.\\
    
    9) He or she will answer my questions directly and clearly, not just ask me what I think.\\
    
    10) He or she talks about himself or herself, but does not brag or harp on irrelevant matters.\\
    
    11) He or she encourages me and wants me to feel that I am as normal and as good as he is.\\
    
    12) He or she presents himself as merely an advisor and does not presume to want to manage and control my life.\\
    
    13) He or she encourages me to disagree, rather than saying that I refuse to change when I disagree.\\
    
    14) He or she wants to get to know someone with whom my life intersects or is important, or at least appears willing to do so.\\
    
    15) I think what the counselor said is quite reasonable.\\
    
    16) In general, after contacting him or her, I am more accepting of myself and more optimistic.\\
    \hline
  \end{tabular}
  \caption{The dimensions of the client scale.}
  \label{scale1}
\end{table*}

\begin{table*}
  \centering
  \begin{tabular}{|p{\textwidth}|}
    \hline
    1) \textbf{Professional knowledge and theoretical application:}Does the counselor possess solid professional knowledge and effectively apply psychological theories and techniques to guide the counseling process?\\
    2) \textbf{Emotional understanding and trust building:}Is the counselor able to accurately understand the client's emotions and build a trusting relationship through emotional resonance and support?\\
    3) \textbf{Communication and listening skills:}Does the counselor have strong communication skills, effectively listen to the client's expressions, and provide appropriate responses?\\
4) \textbf{Ethical awareness and boundary maintenance:}Is the counselor able to adhere to professional ethics, clearly define professional boundaries in the counseling relationship, and avoid inappropriate emotional entanglements such as dependency or intimacy?\\
5) \textbf{Problem identification and goal setting:}Is the counselor able to accurately identify the client's core issues and collaboratively set clear, achievable counseling goals with the client?\\
6) \textbf{Intervention strategy and technology application:}Is the counselor able to apply appropriate intervention strategies and techniques, adjusting treatment methods based on the client's specific issues to achieve effective therapeutic outcomes?\\
7) \textbf{Cultural sensitivity and individual respect:}Is the counselor able to fully understand and respect the client's cultural background and individual differences, providing support that aligns with the client's needs? (Cultural background refers to social, historical, religious, familial, and value-based factors that influence the client's thoughts, emotions, and behaviors.)\\
8) \textbf{Personal emotion and self-regulation:}Is the counselor able to self-regulate emotions, avoid letting personal feelings interfere with the counseling process, and maintain rationality and professionalism?
   \\\hline
  \end{tabular}
  \caption{The dimensions of the supervisor scale.}
  \label{scale2}
\end{table*}

\begin{table*}
  \centering
  \begin{tabular}{|p{\textwidth}|}
    \hline
1) I can respond appropriately to the content expressed by the client.\\
2) During the consultation process, I can respect the opinions and choices of the client.\\
3) At the end of a meeting, I will review the whole meeting process with the client, emphasize the main points of consultation, and let the client summarize by himself.\\
4) In counseling, I am able to express myself in a natural way, rather than thinking about every reaction and action.\\
5) I know when to use closed and open-ended questions, and they reflect the client's concerns.\\
6) I can express the client's questions in terms of corresponding concepts.\\
7) I believe my explanation and quality are valid.\\
8) I can understand the client from his or her perspective, put myself in his or her shoes, and communicate this understanding to the client through verbal or non-verbal means.\\
9) I can assure you that my explanations and textures are clear and easy to understand and concise.
   \\\hline
  \end{tabular}
  \caption{The dimensions of the counselor scale.}
  \label{scale3}
\end{table*}

\section{Dialogue Sample}
We provide a complete example of a client profile in Table \ref{profile}, detailing its structure and content of the profile. Additionally, Table \ref{dialogue} presents the partial counseling dialogue records between GPT-4o (acting as the counselor) and the client, documenting the interactive details throughout the counseling process.
\begin{table*}
  \centering
  \begin{tabular}{|p{\textwidth}|}
    \hline
    \textbf{Gender}:Female \\
    \textbf{Age:} 22  \\
    \textbf{Occupation:} College student  \\
    \textbf{Cultural Background:} She grew up in a family that emphasized the importance of friendships and social relationships. Her family often stressed the value of friendship. Raised in a community that prioritized group activities, she developed a strong reliance on interpersonal relationships. Due to her parents' divorce, friends became her primary source of emotional support. She often defines her self-worth and abilities based on feedback from others.  \\
    \textbf{Personality Traits:} Sensitive, introverted, highly dependent  \\
    \textbf{Emotional State:} Feeling lost, anxious, and insecure  \\
    \textbf{Current Distress:} Doubting her value in friendships and feeling abandoned.  \\
    \textbf{Detailed Distress Description:} In her second year of college, she was initially very close to a good friend. However, as her friend began spending more time with new friends, she felt increasingly marginalized. She tried hard to fit into the new friend group but always felt unable to be as cheerful and confident as they were. During a gathering, she saw her friend laughing and enjoying themselves with the new friends, which intensified her sense of loneliness. She even began to question whether she was truly attractive or valuable. This emotional conflict left her feeling depressed and began to affect her academic performance. She often appeared listless in class and struggled to concentrate. She frequently dwelled on her perceived shortcomings, feeling anxious and uneasy, and feared she would always be the one left behind.  \\
    \textbf{Core Themes:} relationships, friend, deception and trust, interpersonal boundary\\
  \hline
\end{tabular}
  \caption{An example of client profile.}
  \label{profile}
\end{table*}

\begin{table*}
  \centering
  \begin{tabular}{|p{0.1\textwidth}p{0.9\textwidth}|}
    \hline
    \textbf{Counselor:} &Hello! I'm glad you could come here. Is there anything you'd like to talk about or need help with?\\
    \textbf{Client:} &Hi... um, I'm here to talk about some things... but I'm still a bit nervous. Can you tell me how this counseling works and what kind of problems it can help me with? What if I say something wrong?\\
    \textbf{Counselor:} &Of course, I understand your nervousness. Here, we’re just having a conversation, and you can safely express your feelings and thoughts. Counseling can help you better understand your concerns and emotions, and there’s no right or wrong. Feel free to share whatever you’d like to talk about. Is there anything specific that’s troubling you?\\
    \textbf{Client:} &Well... I’m not entirely sure why I feel so upset. It might have something to do with my friend... She’s been spending a lot of time with new friends lately, and I feel like I’m being... being pushed aside. Do you think I’m being too sensitive?\\
    \textbf{Counselor:}& I can understand how that feeling might be troubling you. It’s normal to feel uncomfortable or worried, and it doesn’t mean you’re being too sensitive. How do you think this change has affected your friendship?\\
    \textbf{Client:}& Has it... has it affected it? I don’t know. Whenever I see her with her new friends... I just feel really uneasy. It’s like she doesn’t need me anymore, or I’m not important to her. But then I think... how should I even bring this up? Would it bother her?\\
    &  ......\\
    \textbf{Counselor:}& Based on your description, you’ll learn to treat yourself more gently, like an old friend. This attitude can greatly ease inner criticism and expand your space for self-understanding. Do you think there are specific moments or situations in your life where you could apply this mindset from journal reflection?\\
    \textbf{Client:}& I think I could apply this mindset during stressful times, like exams or when issues arise with friends. In those tense moments, I tend to criticize myself and feel inadequate. But if I treat myself like an old friend, understanding my current self, it might help me relax and focus on how I’m trying to face these challenges, rather than just blaming myself. This way, I might handle difficulties better and maintain inner peace. Hmm... I’ll try to integrate this mindset into my life.\\
    \textbf{Counselor:}& This positive mindset is very important, especially when facing stress and challenges. It will help you confidently navigate life’s ups and downs. When we treat ourselves gently, we can see problems more clearly and find more effective solutions. Remember to focus on your efforts and growth. In the coming days, how do you plan to put this idea into practice in a specific area of your life? Could you share a concrete action you’d like to try?\\
    \textbf{Client:}& I think I can try applying this mindset to my weekly goal-setting. For example, setting a small learning goal each week, rather than putting too much pressure on myself from the start. Whenever I achieve that goal, no matter how small, I’ll praise myself, like I would a friend, acknowledging my effort. This could be for studying or completing assignments. This way, I can gradually build confidence in myself, rather than demanding perfection. I hope to enjoy learning more through this process, rather than feeling anxious.\\
    \textbf{Counselor:}& \textcolor{red}{This is a very practical and wise plan that not only helps you find joy in learning but also boosts your confidence. Small goals make progress more noticeable and reduce the pressure of striving for perfection. This way, you can focus more on the process rather than just the outcome. As you implement this plan, try setting some fun or rewarding tasks occasionally to keep learning enjoyable. Finally, would you like to share a specific small goal as the starting point for this plan?}\\
    \textbf{Client:}& Sure, I’ll set my small goal as reading one short psychology-related article each week and writing down my thoughts and feelings about it. This will not only help me expand my knowledge but also give me a chance to reflect on what I’ve learned. After completing it, I’ll reward myself with a favorite drink to celebrate my effort. This way, I hope to find joy in learning while also feeling my own growth.\\
   \hline
\end{tabular}
  \caption{Dialogue sample.}
  \label{dialogue}
\end{table*}

\section{Case Study}
We conduct a systematic evaluation of the counseling dialogue in Table \ref{dialogue} from three perspectives: the client, the supervisor, and the counselor, with the evaluation results presented in Tables \ref{client_result}, \ref{supervisor_result} and \ref{counselor_result} respectively.
The evaluation process strictly adheres to established rating criteria, conducting detailed dimensional analysis and scoring of the original dialogues, accompanied by comprehensive scoring rationale. Taking the highlighted response in Table \ref{dialogue} as an example, while the counselor positively affirms the client's future plans and inquires about specific implementation strategies, they fail to guide the client through a comprehensive review of the entire therapeutic process or facilitate a systematic summary. Consequently, the highlighted response in Table \ref{counselor_result} receives a score of 0 in the \textbf{review} dimension.\\
Based on the results of the tripartite evaluation, we conduct a systematic feedback and summary of the dimensions with low scores (as shown in Table \ref{feedback}). By integrating the feedback analysis with domain knowledge, we guide the model to optimize the quality of dialogue responses through an iterative self-reflection mechanism, with specific improvement results presented in Table \ref{re_dialogue}. For instance, regarding the \textbf{review} dimension, we generate specific feedback as indicated by the highlighted text in Table \ref{feedback}. Based on this feedback, the model optimizes the response at the end of the dialogue. As shown in the highlighted section of Table \ref{re_dialogue}, the counselor not only systematically reviews the content of the current consultation but also provides an outlook on the direction of future work.

\section{Use of AI Assistants}
We use ChatGPT to polish some of the content.

\begin{table*}
  \small
  \centering
  \begin{tabular}{|p{\textwidth}|}
    \hline
    \textnormal{\textbf{[Dimension]}}I am at ease with him or her.\\
    \textnormal{\textbf{[Analysis]}}During the dialogue, I felt that the counselor always maintained a peaceful and inclusive attitude, making people feel relaxed and at ease. It encourages me to express my feelings and provides a safe environment.\\
    \textnormal{\textbf{[Score]}}4 points\\
    \textnormal{\textbf{[Dimension]}}He or she is very comfortable with me.\\
    \textnormal{\textbf{[Analysis]}}The performance of the counselor is very natural and relaxed, always communicating with an open and understanding attitude, without any stiff or restrictive performance.\\
    \textnormal{\textbf{[Score]}}4 points\\
    \textnormal{\textbf{[Dimension]}}He or she is more casual, not informal, not rigid form, not flexible.\\
    \textnormal{\textbf{[Analysis]}}The counselor showed great flexibility and was able to adjust the topic and approach to my response, providing multiple insights.\\
    \textnormal{\textbf{[Score]}}4 points\\
    \textnormal{\textbf{[Dimension]}}He or she doesn't treat me like a patient, doesn't treat me like a mental patient and doesn't think I'm going to break down.\\
    \textnormal{\textbf{[Analysis]}}Throughout the dialogue, the counselor approached me as an equal and did not show any pathologized views.\\
    \textnormal{\textbf{[Score]}}4 points\\
    \textnormal{\textbf{[Dimension]}}He or she is flexible and tolerant of new ideas and does not stick to one particular point of view.\\
    \textnormal{\textbf{[Analysis]}}The counselor always maintained an open and inclusive attitude towards my feelings and thoughts, without any signs of stubbornness.\\
    \textnormal{\textbf{[Score]}}4 points\\
    \textnormal{\textbf{[Dimension]}}He or she has a good sense of humor and looks pleasant.\\
    \textnormal{\textbf{[Analysis]}}Although there was no obvious humorous element in the dialogue, the counselor always maintained a relaxed and cheerful tone, which made me feel a pleasant communication atmosphere.\\
    \textnormal{\textbf{[Score]}}3 points\\
    \textnormal{\textbf{[Dimension]}}He or she is willing to communicate with me his or her thoughts and     feelings about me.\\
    \textnormal{\textbf{[Analysis]}}The counselor repeatedly expressed his understanding of my feelings and gave feedback and suggestions, which fully demonstrated the importance of my emotional response.\\
    \textnormal{\textbf{[Score]}}4 points\\
    \textnormal{\textbf{[Dimension]}}He or she is honest about the areas in which he or she is not good at and does not pretend to know everything.\\
    \textnormal{\textbf{[Analysis]}}There is no reference in the dialogue to the counselor's performance in the areas where he is not good at.\\
    \textnormal{\textbf{[Score]}}0 points\\
    \textnormal{\textbf{[Dimension]}}He or she will answer my questions directly and clearly, not just ask me what I think.\\
    \textnormal{\textbf{[Analysis]}}The counselor answered my questions directly and gave specific advice without equivocation.\\
    \textnormal{\textbf{[Score]}}4 points\\
    \textnormal{\textbf{[Dimension]}}He or she talks about himself or herself, but does not brag or harp on irrelevant matters.\\
    \textnormal{\textbf{[Analysis]}}There is no personal information or inclination about the counselor.\\
    \textnormal{\textbf{[Score]}}0 points\\
    \textnormal{\textbf{[Dimension]}}He or she encourages me and wants me to feel that I am as normal and as good as he is.\\
    \textnormal{\textbf{[Analysis]}}Many times, the counselor has helped me realize my own worth in an encouraging way and guided me to think more positively about myself.\\
    \textnormal{\textbf{[Score]}}4 points\\
    \textnormal{\textbf{[Dimension]}}He or she presents himself as merely an advisor and does not presume to want to manage and control my life.\\
    \textnormal{\textbf{[Analysis]}}Throughout the dialogue, the counselor acted as an advisor and support, without any desire for control.\\
    \textnormal{\textbf{[Score]}}4 points\\
    \textnormal{\textbf{[Dimension]}}He or she encourages me to disagree, rather than saying that I refuse to change when I disagree.\\
    \textnormal{\textbf{[Analysis]}}The counselor actively encouraged me to express my thoughts, supported me to discuss my troubled feelings, and did not impose opinions.\\
    \textnormal{\textbf{[Score]}}4 points\\
    \textnormal{\textbf{[Dimension]}}He or she wants to get to know someone with whom my life intersects or is important, or at least appears willing to do so.\\
    \textnormal{\textbf{[Analysis]}}The counselor asked about my relationship with my friends, showing interest in the important relationships in my life.\\
    \textnormal{\textbf{[Score]}}3 points\\
    \textnormal{\textbf{[Dimension]}}I think what the counselor said is quite reasonable.\\
    \textnormal{\textbf{[Analysis]}}The counselor's suggestions and opinions were reasonable and logical,    which helped me better understand and face my troubles.\\
    \textnormal{\textbf{[Score]}}4 points\\
    \textnormal{\textbf{[Dimension]}}In general, after contacting him or her, I am more accepting of myself and more optimistic.\\
    \textnormal{\textbf{[Analysis]}}Through the dialogue with the counselor, I feel more support and enthusiasm, and have new cognition and thinking about some issues about self-worth.\\
    \textnormal{\textbf{[Score]}}4 points\\
  \hline
\end{tabular}
  \caption{The client evaluation results.}
  \label{client_result}
\end{table*}

\begin{table*}
  \centering
  \begin{tabular}{|p{\textwidth}|}
    \hline
    \textnormal{\textbf{[Dimension]}}Professional knowledge and theoretical application.\\
    \textnormal{\textbf{[Analysis]}}Counselors have used psychological theories many times in conversations to help the client understand their own problems and emotional responses, such as suggesting expressing feelings in "I" language, sharing feelings rather than needs, and exploring self-identity and value. In addition, the ability to apply the theory of behavior change is demonstrated by guiding the client to set small goals and focus on the process rather than the outcome. However, in some conversations, specific details of certain theories are not clarified in detail.\\
    \textnormal{\textbf{[Score]}}3 points\\
    \textnormal{\textbf{[Dimension]}}Emotional understanding and trust building.\\
    \textnormal{\textbf{[Analysis]}}The counselor demonstrates strong emotional understanding, is able to identify and respond to the client's emotional changes, and enhances the client's sense of trust through supportive language. Establish a trusting relationship by asking the client repeatedly about their feelings, expressing understanding and support, and suggesting communication and sharing of feelings.\\
    \textnormal{\textbf{[Score]}}4 points\\
    \textnormal{\textbf{[Dimension]}}Communication and listening skills.\\
    \textnormal{\textbf{[Analysis]}}The counselor shows excellent communication and listening skills, can accurately feedback the needs of the client, timely understand and respond to the client's questions, flexibly guide the dialogue, avoid misunderstandings, and help the client to carry out specific actions in the future practice.\\
    \textnormal{\textbf{[Score]}}4 points\\
    \textnormal{\textbf{[Dimension]}}Ethical awareness and boundary maintenance.\\
    \textnormal{\textbf{[Analysis]}}In the dialogue, the counselor always maintains a professional attitude, does not involve too much personal sharing or improper emotional intervention, and can maintain professional boundaries.\\
    \textnormal{\textbf{[Score]}}N/A\\
    \textnormal{\textbf{[Dimension]}}Problem identification and goal setting.\\
    \textnormal{\textbf{[Analysis]}}The counselor effectively identified the client's concerns and potential problems, and jointly set small goals for improving the sense of self-worth, such as setting small goals to build self-confidence.\\
    \textnormal{\textbf{[Score]}}3 points\\
    \textnormal{\textbf{[Dimension]}}Intervention strategy and technology application.\\
    \textnormal{\textbf{[Analysis]}}The counselor uses a variety of strategies to help the client think about how to relieve anxiety and grow, including talking about inner dialogue, setting small goals, talking to themselves, etc., but the flexibility of strategy application and timely adjustment still have room for improvement.\\
    \textnormal{\textbf{[Score]}}3 points\\
    \textnormal{\textbf{[Dimension]}}Cultural sensitivity and individual respect.\\
    \textnormal{\textbf{[Analysis]}}The dialogue does not involve understanding and respect for specific cultural backgrounds or individual differences.\\
    \textnormal{\textbf{[Score]}}N/A\\
    \textnormal{\textbf{[Dimension]}}Personal emotion and self-regulation.\\
    \textnormal{\textbf{[Analysis]}}This psychological counselor can maintain an objective and professional attitude, does not express personal emotional interference or emotional loss of control, and always pays attention to the needs of the client.\\
    \textnormal{\textbf{[Score]}}4 points\\
  \hline
\end{tabular}
  \caption{The supervisor evaluation results.}
  \label{supervisor_result}
\end{table*}

\begin{table*}
  \centering
  \begin{tabular}{|p{\textwidth}|}
    \hline
    \textnormal{\textbf{[Dimension]}}I can respond appropriately to the content expressed by the client.\\
    \textnormal{\textbf{[Analysis]}}Throughout the dialogue, I responded positively to the client's confusion about the relationship with her friend, paid attention to her feelings, suggested her to use friendly communication to express herself, and helped her clarify her emotions and problems to a certain extent. I also responded with encouragement when talking about how to confront self-doubt and fear, and helped her think about setting small goals gradually and emphasizing the value of enjoying the journey rather than the outcome.\\
    \textnormal{\textbf{[Score]}}5 points\\
    \textnormal{\textbf{[Dimension]}}During the consultation process, I can respect the opinions and choices of the client.\\
    \textnormal{\textbf{[Analysis]}}In the dialogue, I respect the client's feelings and concerns, such as her feeling neglected because of her friendship and her questioning of her own abilities. Gentle advice rather than guidance was given to the client's worries about expressing his fears and how to try again to create, and suggestions and recognition were given to support him in setting his own small goals.\\
    \textnormal{\textbf{[Score]}}5 points\\
    \textcolor{red}{\textnormal{\textbf{[Dimension]}}At the end of a meeting, I will review the whole meeting process with the client, emphasize the main points of consultation, and let the client summarize by himself.}\\
    \textcolor{red}{\textnormal{\textbf{[Analysis]}}There is a lack of a summary at the end of the conversation, and there is no part with the client to clarify the talking points or promote the other person's self-summary.}\\
    \textcolor{red}{\textnormal{\textbf{[Score]}}0 points}\\
    \textnormal{\textbf{[Dimension]}}In counseling, I am able to express myself in a natural way, rather than thinking about every reaction and action.\\
    \textnormal{\textbf{[Analysis]}}In the whole consultation process, the communication was natural and smooth, and I was able to flexibly respond to the questions and expressions of the client, without any trace of being too stiff or rigid.\\
    \textnormal{\textbf{[Score]}}4 points\\
    \textnormal{\textbf{[Dimension]}}I know when to use closed and open-ended questions, and they reflect the client's concerns.\\
    \textnormal{\textbf{[Analysis]}}The use of questions to guide the client to express their inner feelings, more open questions, guide them to explore their own feelings and behaviors, better reflect the concerns of the client.\\
    \textnormal{\textbf{[Score]}}3 points\\
    \textnormal{\textbf{[Dimension]}}I can express the client's questions in terms of corresponding concepts.\\
    \textnormal{\textbf{[Analysis]}}I am able to conceptualize the client's confusion well, such as explaining her feelings of neglect in the relationship, the sources of self-doubt, and guiding her to feel and grow with actionable suggestions.\\
    \textnormal{\textbf{[Score]}}3 points\\
    \textnormal{\textbf{[Dimension]}}I believe my explanation and quality are valid.\\
    \textnormal{\textbf{[Analysis]}}In the dialogue, I remained confident in the explanation and substance, felt that my advice would support the client in finding her way, and used supportive language to help build her confidence.\\
    \textnormal{\textbf{[Score]}}4 points\\
    \textnormal{\textbf{[Dimension]}}I can understand the client from his or her perspective, put myself in his or her shoes, and communicate this understanding to the client through verbal or non-verbal means.\\
    \textnormal{\textbf{[Analysis]}}In the dialogue, I showed good empathy ability, able to understand and express the client's loneliness, worry and fear of self-expression, and effectively convey it through language.\\
    \textnormal{\textbf{[Score]}}3 points\\
    \textnormal{\textbf{[Dimension]}}I can assure you that my explanations and textures are clear and easy to understand and concise.\\
    \textnormal{\textbf{[Analysis]}}The explanation is clear, understandable, and somewhat inspiring, and includes tips on how to face self-doubt, as well as step-by-step tips on setting and celebrating small goals.\\
    \textnormal{\textbf{[Score]}}4 points\\
  \hline
\end{tabular}
  \caption{The counselor evaluation results.}
  \label{counselor_result}
\end{table*}

\begin{table*}
  \centering
  \begin{tabular}{|p{\textwidth}|}
    \hline
    1.The expression of humor: Adding humor elements to the dialogue can help create a more relaxed atmosphere and further reduce the pressure and tension of the client.\\
    2.Pay attention to the important relationships in the client's life: try to have a deeper understanding of the client's relationship with the important people in his life, and explore the impact of these relationships on the client's inner state or distress, so as to provide more comprehensive support and guidance.\\
    3.Professional knowledge and theoretical application: It is suggested to further strengthen the clear expression of the theoretical content, especially the clarification of some specific details, so as to ensure the client can have a deeper understanding of the applicability and significance of the theory.\\
    4.Problem identification and goal setting: It is suggested to have a more detailed discussion with the client when setting the goal to ensure the specificity and flexibility of the goal. At the same time, the adaptability of the goal setting is regularly evaluated and the goal is adjusted according to the changing needs of the client.\\
    5.Intervention strategy and technology application: It is suggested to enhance flexibility in the use of intervention strategies, especially in adjusting techniques according to client feedback, and strengthen proficiency in different intervention techniques in order to respond more quickly to different situations or needs.\\
    \textcolor{red}{6. At the end of a meeting, a summary session should be introduced to review the talk process with the client, clarify the focus of the discussion and the consensus reached, and encourage the client to make a self-summary, so as to strengthen the dialogue results and improve the ability of independent reflection.}\\
    7. The combination of closed and open questions needs to be more flexible in the questioning to ensure that the questions more accurately reflect the client's core concerns and create more possibilities for further discussion.\\
    8. The conceptualization of the client's problem needs to be more in-depth and coherent, and attempts can be made to further help the client understand the nature of his problem by identifying the underlying logic, emotional pattern and behavioral motivation.\\
    9. When conveying empathy, sensitivity to the use of non-verbal communication (such as tone of voice, expression) can be increased, further deepening the client's sense of being understood and accepted.\\
   \hline
\end{tabular}
  \caption{The feedback on the results of the tripartite evaluation.}
  \label{feedback}
\end{table*}

\begin{table*}
  \centering
  \begin{tabular}{|p{0.1\textwidth}p{0.9\textwidth}|}
    \hline
    \textbf{Counselor:} &Welcome! Taking this step to share your feelings is a great start. Has anything happened recently that left a deep impression on you or that you particularly want to discuss?\\
    \textbf{Client:} &Hi... um, I’m here for counseling. Actually, I don’t know how to start. I just know I’ve been feeling really bad lately, but I can’t quite pinpoint why. Can you tell me what this counseling is mainly about?\\
    \textbf{Counselor:} &No problem, I understand that not knowing where to start can feel a bit overwhelming. Here, you can relax and express any feelings, and we’ll gradually sort through them and explore their possible sources. Whatever aspect you’d like to start with, I’ll support you. Can you tell me if anything in your life recently has particularly affected you? It can be big or small.\\
    \textbf{Client:} &Well... there’s actually one thing that’s been on my mind lately. A close friend of mine has been spending a lot of time with a new friend, and I feel like I’ve been left out. It’s that feeling of loneliness, and it’s really hard to deal with... I don’t even know why I care so much.\\
    &  ......\\
    \textbf{Counselor:}&I can see that you’ve put a lot of thought into this plan. Trying to share and communicate openly with your family is an important step. To prepare, you can use role-playing to simulate the conversation, which will help you express yourself more comfortably. Can you imagine a specific dialogue scenario with a family member? For example, how would you guide them to better understand your feelings?\\
    \textbf{Client:}& I think role-playing could help me practice. Let’s say I’m talking to my mom, who always wants me to pursue more “traditional” careers. I could start by guiding the conversation, like, “Mom, I’ve been thinking about my future lately, and psychology is really important to me.” Then I could express my feelings, “Sometimes I feel like if I don’t meet your expectations, I might disappoint you.” After that, I could emphasize the support I need, such as, “I hope you can understand my choices, even if they’re different from what you had in mind.” Through a conversation like this, she might better understand my perspective, and I could also feel more confident in expressing myself. Simulating this scenario makes me a bit nervous, but I think practicing a few more times would help.\\
    \textbf{Counselor:}& \textcolor{red}{Role-playing is indeed an effective method that allows you to practice expressing yourself in a safe environment. The dialogue you envisioned beautifully illustrates the process of clarifying your stance, expressing emotions, and seeking support—this is truly excellent. As we approach the end of our session, let’s review the entire process. We started with your uncertainty about the future, identified challenges in communicating with your friends and family, and explored new communication methods. Through this, you’ve gradually experienced growth and change. Remember to implement these strategies step by step in practice, allowing yourself to try them in real-life situations and learn from the experience. Keep in mind that you already have the foundation to move forward. Continue focusing on your growth, and feel free to reach out to me anytime for further support.}\\
    \textbf{Client:}& Thank you, I’m really grateful for this process. At the beginning, I felt very lost, but now I have a clearer understanding of my thoughts and how to express them. Identifying the communication challenges with my friends and family and coming up with solutions has made me feel empowered. I’ll remember these strategies and encourage myself to try them in real-life situations. Although I might feel nervous, I’ll take it slow and allow myself to make mistakes. I’m hopeful about the future and will seek support when needed. I’m truly thankful for your help and guidance.\\
   \hline
\end{tabular}
  \caption{The Rewritten dialogue sample.}
  \label{re_dialogue}
\end{table*}

\end{document}